\pdfoutput=1
\documentclass[10pt,journal,compsoc]{IEEEtran}

\usepackage{amsmath,amssymb}
\usepackage{graphicx}
\usepackage{booktabs}
\usepackage{xcolor}
\usepackage{microtype}
\usepackage{xspace}
\usepackage{multirow}
\usepackage{colortbl}
\usepackage{makecell}
\usepackage{bm}
\usepackage{mathtools}
\usepackage{array}
\usepackage{pifont}
\usepackage{url}
\usepackage[shortlabels]{enumitem}
\ifCLASSOPTIONcompsoc
  \usepackage[nocompress]{cite}
\else
  \usepackage{cite}
\fi

\newcommand{\method}{WACO\xspace}
\newcommand{\bench}{LIBERO-PoseShift\xspace}
\newcommand{\pizf}{$\pi_{0.5}$\xspace}
\newcommand{\pizero}{$\pi_0$\xspace}
\newcommand{\gain}[1]{{\scriptsize\color{red}#1}}

\newcommand{\yes}{{\color{green!60!black}\ding{51}}}
\newcommand{\no}{{\color{red!80!black}\ding{55}}}
\newcommand{\na}{--}

\DeclareMathOperator{\Exp}{Exp}

\DeclareMathOperator{\FK}{FK}
\DeclareMathOperator{\SE}{SE}

\newcommand{\T}[1]{\mathbf{T}_{#1}}
\newcommand{\Tbc}{\mathbf{T}_{bc}}
\newcommand{\Tbcz}{\mathbf{T}^{0}_{bc}}
\newcommand{\Twb}{\mathbf{T}_{wb}}
\newcommand{\Twbz}{\mathbf{T}^{0}_{wb}}
\newcommand{\Twc}{\mathbf{T}_{wc}}
\newcommand{\Twcz}{\mathbf{T}^{0}_{wc}}
\newcommand{\dTc}{\Delta\mathbf{T}_{c}}
\newcommand{\dTb}{\Delta\mathbf{T}_{b}}
\newcommand{\hdTc}{\widehat{\Delta\mathbf{T}}_{c}}
\newcommand{\hdTb}{\widehat{\Delta\mathbf{T}}_{b}}
\newcommand{\hTbc}{\widehat{\mathbf{T}}_{bc}}
\newcommand{\vect}[1]{\mathbf{#1}}

\newcommand{\Cmat}{\mathbf{C}}

\newcolumntype{L}[1]{>{\raggedright\arraybackslash}p{#1}}
\newcolumntype{C}[1]{>{\centering\arraybackslash}p{#1}}
\definecolor{rowgray}{gray}{0.93}
\definecolor{meanblue}{RGB}{232,241,251}
\definecolor{oursblue}{RGB}{220,235,250}

\renewcommand{\paragraph}[1]{\vspace{.35em}\noindent\textbf{#1}}

\definecolor{linkblue}{rgb}{0.21,0.49,0.74}
\usepackage[breaklinks,colorlinks,allcolors=linkblue]{hyperref}
\usepackage[capitalize]{cleveref}
\crefname{figure}{Fig.}{Figs.}
\Crefname{figure}{Fig.}{Figs.}
\crefname{table}{Table}{Tables}
\Crefname{table}{Table}{Tables}
\crefname{section}{Section}{Sections}
\Crefname{section}{Section}{Sections}
\crefname{subsection}{Section}{Sections}
\Crefname{subsection}{Section}{Sections}
\crefname{equation}{Eq.}{Eqs.}
\Crefname{equation}{Eq.}{Eqs.}

\newcommand{\emailaddr}[1]{\begingroup\urlstyle{same}\nolinkurl{#1}\endgroup}

\begin{document}

\title{Towards Pose-Robust Few-Shot\\Vision-Language-Action Models}

\author{Wenhao~Li, Zhongyi~Han, Yuhao~Wei, Liang~Han, Yilong~Yin, and~Liqiang~Nie%
\thanks{\upshape\hspace*{1em}Wenhao Li, Zhongyi Han and Yilong Yin are with the School of Software, Shandong University, Jinan 250101, China (E-mail: \protect\emailaddr{wenhao.li@mail.sdu.edu.cn}; \protect\emailaddr{zhongyi.han@sdu.edu.cn}; \protect\emailaddr{ylyin@sdu.edu.cn}).\protect\\
\hspace*{1em}Wenhao Li and Yuhao Wei are with Shenzhen Loop Area Institute, Shenzhen 518038, China (E-mail: \protect\emailaddr{yuhaowei@slai.edu.cn}).\protect\\
\hspace*{1em}Yuhao Wei is also with the School of Life Science and Technology, Xidian University, Xi'an 710126, China.\protect\\
\hspace*{1em}Liang Han and Liqiang Nie are with the School of Computer Science and Technology, Harbin Institute of Technology (Shenzhen), 518055, China (E-mail: \protect\emailaddr{han.liang@hit.edu.cn}; \protect\emailaddr{nieliqiang@gmail.com}).\protect\\
\hspace*{1em}(Corresponding author: Liqiang Nie.)}}

\markboth{}{}

\IEEEtitleabstractindextext{%
\begin{abstract}
Fine-tuning a vision-language-action (VLA) model on a few dozen demonstrations is the standard way to put a manipulation policy on a robot, yet the resulting policy silently assumes that the camera and the robot base stay exactly where they were during data collection: a bumped tripod or a base re-docked a few centimetres away makes it reach for the wrong place.
We show that both failures are one phenomenon, a change of the camera-to-base extrinsic that the policy has memorised, and that the robot carries everything needed to undo it.
We introduce \method, a world-anchored canonicalization framework for VLA models: the arm, placed by forward kinematics at the measured joint angles, serves as a metric calibration target, and the static scene serves as the world anchor that tells a moved camera from a moved robot.
From a single RGB-D frame, \method estimates both shifts without ground-truth extrinsics or extra data, canonicalises the image to the calibration viewpoint and the state, action and start pose to the calibration base frame, and fine-tunes the unmodified policy under the artefacts of its own estimator.
We also introduce \bench, the first benchmark for RGB VLAs with graded camera, base and compound shifts.
Across \bench, the LIBERO-Plus camera track and four real-world tasks on a UR5e robot, \method consistently outperforms the standard few-shot recipe and prior viewpoint-robust methods, raising shifted success by 13.2 points in simulation and by 36.7 points on the real robot, recovering 90\% of the simulation gap to a ground-truth oracle and 87\% of the real-robot gap, and establishing a new state of the art among methods that use no perturbed training data, with no test-time calibration.
Project page: \url{https://peacelwh.github.io/WACO/}.
\end{abstract}

\begin{IEEEkeywords}
Vision-language-action models, few-shot imitation learning, robot manipulation, viewpoint robustness, camera-to-robot calibration, canonicalization, benchmark.
\end{IEEEkeywords}}

\maketitle
\IEEEdisplaynontitleabstractindextext
\IEEEpeerreviewmaketitle

\begin{figure*}[t]
\centering
\includegraphics[width=\linewidth]{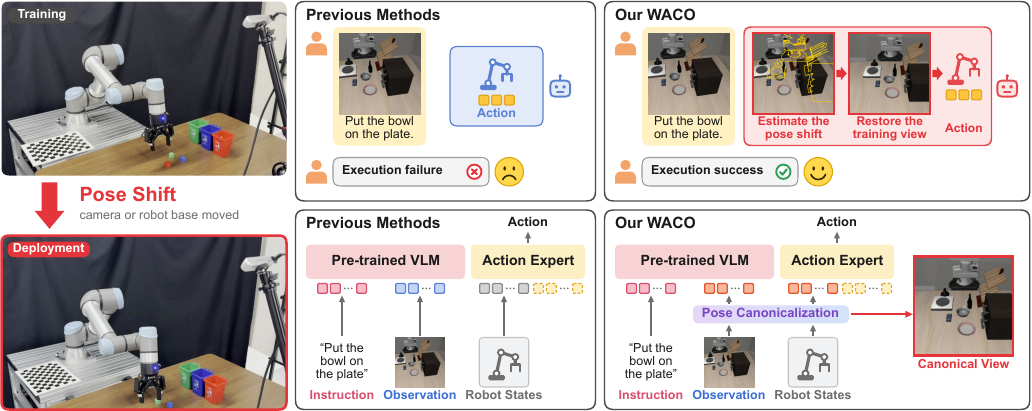}
\caption{\textbf{Overview.}
A few-shot fine-tuned VLA memorises the camera-to-base extrinsic of its demonstrations and misses the target once the camera is bumped or the robot is re-docked (left: our UR5e setup at the calibration pose and after the table and camera were moved 10\,cm away from the robot).
\method estimates both shifts from a single RGB-D frame, with the arm placed by forward kinematics as the metric anchor and the static scene as the world anchor, and canonicalises the observation back to the calibration viewpoint and base frame, so that an unmodified \pizf sees the world it was trained in (right; yellow: outline of the calibration reference in the live view).
We evaluate on \bench, the LIBERO-Plus camera track and four real-world UR5e tasks.}
\label{fig:teaser}
\end{figure*}

\IEEEraisesectionheading{\section{Introduction}\label{sec:intro}}

\IEEEPARstart{R}{ecent} vision-language-action (VLA) models~\cite{brohan2023rt2,kim2024openvla,black2024pi0,pi2025pi05} have made few-shot fine-tuning the standard route to a deployed manipulation policy: mount a camera, dock the robot at a table, teleoperate 50--100 episodes, fine-tune a pre-trained model and deploy.
In distribution the recipe works remarkably well; the public LIBERO release of \pizf succeeds in $97\%$ of episodes~\cite{cadene2024lerobot,liu2023libero}.

However, the demonstrations are collected from a single camera pose and a single robot-base pose, and the policy silently bakes both into its weights.
In practice the tripod gets bumped, the camera is re-mounted after maintenance, or the mobile base re-docks a few centimetres off; each of these is a rigid \emph{pose shift} of a few centimetres and degrees, and each makes the policy reach for the wrong place (\cref{fig:teaser}).
A $5^\circ$ camera yaw moves the target by 21 pixels in a $256^2$ image; a 5\,cm base offset moves it by 5\,cm in the frame in which the policy commands its end effector, while the third-person image barely changes.
LIBERO-Plus~\cite{fei2026liberoplus} has quantified the camera half of this problem, with \pizero dropping from $94\%$ to $16\%$ under camera perturbations, but the base half has neither a benchmark nor a method for RGB VLAs.
This raises two questions: (1) \emph{Can a policy trained at a single camera and base pose be made robust to pose shifts without data from other poses?} (2) \emph{Can the shifts be recovered at deployment without a calibration object, ground-truth extrinsics or new data?}

Existing remedies answer the first question with data or with geometry.
Data-side methods synthesise thousands of viewpoints~\cite{chen2024roviaug,tian2024vista,yang2025robosplat,fei2026liberoplus} or regularise the policy across simulator-rendered views~\cite{huang2026crossview}; they need perturbed data at a scale that few-shot fine-tuning cannot afford, and they have no counterpart for a moved base, whose action labels change as well as its images.
Geometry-side methods condition the policy on camera rays~\cite{jiang2025camcond,wang2026rayvit,peng2026g3vla} or re-synthesise the training view~\cite{heo2026anycamvla}, but they need the true extrinsics at test time, which a bumped camera no longer has; the one calibration-free VLA we know of~\cite{li2026camvla} regresses the hand-eye transform with an error of $9^\circ$ at a $15^\circ$ shift.
None of them addresses a moved robot base.

We answer both questions with \emph{world-anchored canonicalization}: instead of asking the policy to learn to generalise across poses from data, we recover the shifts from what the robot already knows and map the observation back to the pose it was trained at.
Three observations make this possible.
First, both shifts are one quantity: each changes the camera-to-base extrinsic $\Tbc$, and a base shift also moves the objects in the base frame.
Second, the arm is its own calibration target: forward kinematics at the measured joint angles places every link in the base frame, so the visible arm pins down $\Tbc$ from a single RGB-D frame, without a checkerboard and without training~\cite{lee2020dream,lu2024ctrnetx}.
Third, the arm alone is not enough: every camera and base shift with the same $\Tbc$ renders the arm identically yet demands different actions, the classical $AX{=}YB$ ambiguity~\cite{zhuang1994axyb}, and the static scene provides the missing world anchor for free.
\method (\cref{fig:overview}) turns these observations into a training-free estimator with a ground-truth-free consistency gate, a canonicaliser that warps the image with the camera shift only and re-expresses state, action and start pose with the base shift only, and an error-aware fine-tuning that trains the policy on the canonicaliser's own artefacts.
The policy architecture is untouched, and a base shift is reduced to ``the object is elsewhere on the table'', a variation the demonstrations already cover.

Our contributions are as follows:
\begin{itemize}[leftmargin=*,itemsep=1pt,topsep=2pt]
\item We formulate camera and base shifts of a VLA as one extrinsic change, expose the gauge ambiguity of arm-based self-calibration, and show that LIBERO's world-frame convention silently absorbs base shifts.
\item We propose \method, a calibration-free two-anchor estimator, a channel-separated canonicaliser and an estimator-matched augmentation that make an unmodified \pizf robust to camera, base and compound shifts.
\item We introduce \bench, the first RGB-VLA benchmark with graded base and compound shifts and a base-frame protocol that is isomorphic to a real robot.
\item On \bench, the LIBERO-Plus camera track and four real-world tasks on a UR5e robot, \method outperforms the standard few-shot recipe and prior viewpoint-robust methods, recovers most of the gap to a ground-truth oracle, and gains the most where demonstrations are scarcest.
\end{itemize}

\section{Related Work}
\label{sec:related}

\begin{table}[t]
\centering
\footnotesize
\setlength{\tabcolsep}{3pt}
\caption{\textbf{Related Work Summary.} Which pose shifts a method handles, what it requires beyond the demonstrations, and whether the policy architecture is left unchanged. \emph{Extr.}: ground-truth extrinsics at test time; \emph{Pert.\ data}: perturbed or multi-view training trajectories; \emph{Sim}: a simulator that re-renders states; \emph{Arch.}: policy architecture unchanged; \emph{Calib.-free}: no test-time calibration object or procedure.}
\label{tab:related}
\resizebox{\linewidth}{!}{%
\begin{tabular}{l ccc ccc c c}
\toprule
 & \multicolumn{3}{c}{Shift handled} & \multicolumn{3}{c}{Requires} & & \\
\cmidrule(lr){2-4}\cmidrule(lr){5-7}
Method & Cam. & Base & Both & Extr. & Pert.\ data & Sim & Arch. & Calib.-free \\
\midrule
OpenVLA-OFT+~\cite{fei2026liberoplus} & \yes & \no & \no & \no & \yes & \yes & \yes & \yes \\
RoVi-Aug~\cite{chen2024roviaug} & \yes & \no & \no & \no & \yes & \no & \yes & \yes \\
VISTA~\cite{tian2024vista} & \yes & \no & \no & \no & \yes & \no & \yes & \yes \\
Cross-View AC~\cite{huang2026crossview} & \yes & \no & \no & \no & \no & \yes & \yes & \yes \\
RayViT~\cite{wang2026rayvit} & \yes & \no & \no & \yes & \yes & \no & \no & \no \\
AnyCamVLA~\cite{heo2026anycamvla} & \yes & \no & \no & \yes & \no & \no & \yes & \no \\
OC-VLA~\cite{zhang2025ocvla} & \yes & \no & \no & \yes & \yes & \no & \no & \no \\
CamVLA~\cite{li2026camvla} & \yes & \no & \no & \no & \yes & \no & \no & \yes \\
FTM/FLA~\cite{li2026vlageneralizable} & \yes & \no & \no & \no & \yes & \no & \no & \yes \\
DINOBot~\cite{dipalo2024dinobot} & \no & \yes & \no & \no & \no & \no & \na & \yes \\
\rowcolor{oursblue}
\textbf{Ours} & \yes & \yes & \yes & \no & \no & \no & \yes & \yes \\
\bottomrule
\end{tabular}}
\end{table}

\paragraph{Pose-Shift Vulnerability of VLAs.}
LIBERO-Plus~\cite{fei2026liberoplus} finds camera and robot initial-state perturbations to be the two most damaging of seven factors for \pizero, OpenVLA-OFT and UniVLA~\cite{black2024pi0,kim2025oft,bu2025univla}; LIBERO-PRO~\cite{zhou2025liberopro}, COLOSSEUM~\cite{pumacay2024colosseum}, AGNOSTOS~\cite{zhou2025agnostos} and REALM~\cite{sedlacek2025realm} reach similar conclusions, and Mobi-$\pi$~\cite{yang2025mobipi} shows that a base offset of about 3\,cm halves the success of a fixed-base policy.
None of them perturbs the base of an RGB policy or combines camera and base shifts.
A wrist camera improves viewpoint robustness~\cite{hsu2022hands}, which we confirm and rely on.

\paragraph{Viewpoint Robustness by Data.}
The dominant remedy is to show the policy more viewpoints: simulator re-rendering~\cite{fei2026liberoplus}, novel-view synthesis~\cite{chen2024roviaug,tian2024vista,zhu2024nerfaug,yang2025robosplat,wang2026gensplat,cai2026robonvs,xu2025egodemogen,pan2026actionview}, sim-to-real translation~\cite{coholich2026mango} or large augmented corpora~\cite{ji2025oxeauge}; Cross-View Action Consistency~\cite{huang2026crossview} regularises \pizf across simulator-rendered view pairs of the same state.
These approaches need thousands of extra trajectories or a simulator that re-renders arbitrary states, and they only change the image; when the base moves, the action labels must move with it (\cref{tab:related}).
Our augmentation needs only the demonstrations, and base shifts are handled analytically.

\paragraph{Geometry in the Policy.}
Pl\"ucker-ray conditioning~\cite{jiang2025camcond,wang2026rayvit,peng2026g3vla}, camera-frame actions with hand-eye prediction~\cite{zhang2025ocvla,argus2025cvla,li2026camvla}, robot-centric pointmaps~\cite{lee2026seer} and explicit 3D inputs~\cite{qu2025spatialvla,zhang2026geo3d,goyal2023rvt,shridhar2022peract,ke2024diffuseractor} inject geometry into the policy, but with a single training extrinsic the conditioning signal is constant during training, and after a shift the true extrinsic is unknown.
AnyCamVLA~\cite{heo2026anycamvla} re-synthesises the training view but assumes ground-truth extrinsics, and lightweight adapters~\cite{li2026vlageneralizable} need data from the new viewpoint.
\method supplies the missing extrinsic from the robot itself and leaves the policy untouched.

\paragraph{Calibration and Align-and-Replay.}
DREAM~\cite{lee2020dream}, CtRNet~\cite{lu2023ctrnet,lu2024ctrnetx}, RoboPEPP~\cite{goswami2025robopepp} and RoboTAG~\cite{liu2025robotag} estimate the camera-to-robot pose from one image of the arm, and KineDepth~\cite{atar2024kinedepth} uses the arm to scale monocular depth; all of them yield only $\Tbc$, which \cref{sec:method:pre} shows is insufficient on its own.
Coarse-to-fine imitation~\cite{johns2021coarse}, one-shot pose estimation~\cite{vitiello2023oneshot} and DINOBot~\cite{dipalo2024dinobot} estimate the live-to-demonstration pose and replay a trajectory; \method keeps a closed-loop VLA in the loop, handles both shifts jointly and trains the policy for the estimator's residual error, whereas a mask-based policy drops from 0.98 to 0.64 under an extrinsic error of 1\,cm and $5^\circ$~\cite{lepert2025shadow}.

\section{Method}
\label{sec:method}

\begin{figure*}[t]
\centering
\includegraphics[width=\linewidth]{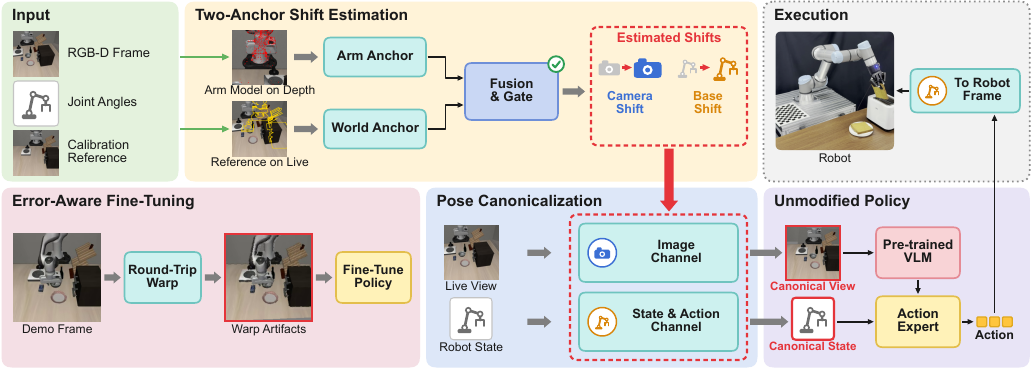}
\caption{\textbf{Model Pipeline.} \method wraps an unmodified \pizf. At reset, a training-free front end registers the FK-placed arm model to the observed depth (\emph{arm anchor}) and matches the static scene to an empty-workspace reference (\emph{world anchor}); their fusion yields $\hdTc$, $\hdTb$ and an out-of-plane residual that gates the estimate. Channel A warps the image with $\hdTc$ only; channel B re-expresses state, action and start pose with $\hdTb$ only. The policy is fine-tuned once on the canonicaliser's own artefacts (bottom left).}
\label{fig:overview}
\end{figure*}

\subsection{Preliminary}
\label{sec:method:pre}

\paragraph{Problem Formulation.}
A VLA policy is fine-tuned on demonstrations collected from a single camera pose and a single robot-base pose, and must keep working after either has moved by a few centimetres and degrees, without new data or calibration.
Let $W$ be the world frame of table, objects and third-person camera, $B$ the robot base frame and $C$ the camera frame; a superscript $0$ marks demonstration time, $\T{ab}$ maps frame $b$ to frame $a$, and the extrinsic $\Tbcz=(\Twbz)^{-1}\Twcz$, measured once before data collection, is the only calibration we assume.
The policy receives a third-person image, a wrist image and a state with the end-effector pose in $B$ and the gripper opening, and outputs end-effector deltas $\vect a_t=(\Delta\vect p,\Delta\bm\omega,g)$ that the controller applies in $B$; the joint angles $\vect q_t$ are measured, and $\FK(\vect q)$ places every link of the arm in $B$.
At deployment the camera and the base may have moved,
\begin{equation}
\Twc=\Twcz\,\dTc,\qquad \Twb=\Twbz\,\dTb,
\label{eq:shift}
\end{equation}
with an unknown rigid camera motion $\dTc\in\SE(3)$ and an unknown planar base motion $\dTb\in\SE(2)$, while the table and the objects stay in place.
Substituting \cref{eq:shift} into $\Tbc=\Twb^{-1}\Twc$ gives
\begin{equation}
\Tbc=\dTb^{-1}\,\Tbcz\,\dTc ,
\label{eq:decomp}
\end{equation}
so \emph{both shifts change the extrinsic} that the policy has memorised (Proposition~1, \cref{sup:derivations}).
A pure camera shift leaves the objects, the correct actions and the wrist view unchanged; a pure base shift also moves every object by $\dTb^{-1}$ in $B$, while the third-person image changes only where the arm is.

\paragraph{Gauge Ambiguity.}
The arm pixels together with $\vect q_t$ determine $\Tbc$ and nothing more: every pair $(\dTc,\dTb)$ that satisfies \cref{eq:decomp} renders the arm identically, yet the pairs demand different actions.
Attributing everything to the camera (\emph{base gauge}) and re-projecting the image is wrong under a base shift, because the object has moved in $B$; attributing everything to the base (\emph{camera gauge}), as camera-frame VLAs implicitly do~\cite{zhang2025ocvla,li2026camvla}, leaves the policy to generalise across viewpoints on its own.
This is the $AX{=}YB$ identifiability problem~\cite{zhuang1994axyb,dornaika1998axyb} in disguise (Proposition~2): separating the two factors of \cref{eq:decomp} needs a second observation anchored in $W$, and with a camera fixed in the world the static scene is that observation.

\paragraph{Protocol Leakage.}
LIBERO expresses state and actions in the \emph{world} frame, which hands a moved robot the correct target for free, since neither the object nor the reported state moves in $W$.
A real robot has no world frame, so we evaluate under a \emph{base-frame} protocol anchored at the calibration pose, in which a moved robot reports the pose it \emph{believes} it has, exactly as hardware does.
The base-frame protocol of \bench anchors state and action in the base pose at calibration time and rotates the policy's deltas from that frame to the world before the operational-space controller, using the true base pose inside the simulator only; in the clean condition it is bit-identical to the original dataset, so no training data has to be converted.

\paragraph{Pipeline.}
\method (\cref{fig:overview}) consists of a training-free \emph{two-anchor estimator} of $(\hdTc,\hdTb)$ (\cref{sec:method:est}), a \emph{channel-separated canonicaliser} (\cref{sec:method:canon}) and an \emph{error-aware fine-tuning} of the unmodified \pizf (\cref{sec:method:train}).
At test time it uses sensor data and the one-off calibration only; ground-truth poses, segmentation and simulator state are never read.

\subsection{Two-Anchor Shift Estimation}
\label{sec:method:est}

\begin{figure*}[t]
\centering
\includegraphics[width=\linewidth]{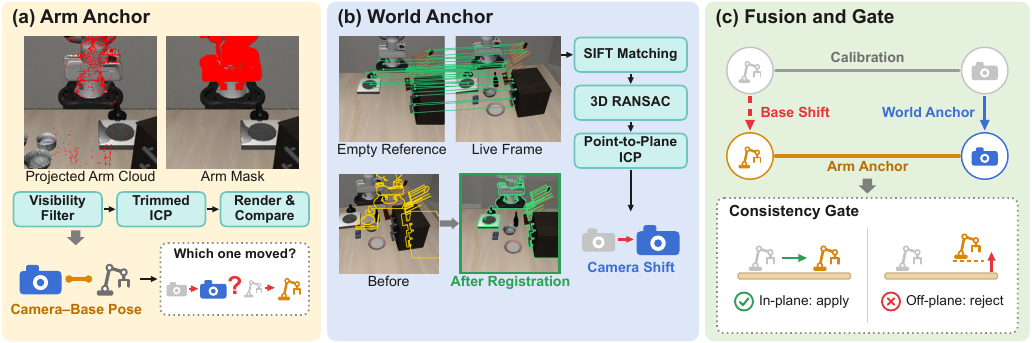}
\caption{\textbf{Two-Anchor Estimation.} \textbf{(a)} Arm anchor: the visible part of the FK-placed arm cloud is registered to the observed depth by trimmed ICP and ranked by a render-and-compare score. \textbf{(b)} World anchor: the live frame is registered to the empty-workspace reference by SIFT, 3D RANSAC and point-to-plane ICP. \textbf{(c)} Fusion closes $\Tbc=\dTb^{-1}\Tbcz\dTc$; the discarded height, roll and pitch of the base shift form a ground-truth-free consistency check.}
\label{fig:estimator}
\end{figure*}

\paragraph{Calibration Bundle.}
Once per workspace we record $\Tbcz$, the nominal base pose $\Twbz$ and one RGB-D frame of the \emph{empty} workspace from the nominal camera (\cref{fig:estimator}b); on a real robot this is a photograph of the bare table taken during setup.
Every point in it is static by construction, so neither segmentation nor a hand-designed background mask is needed.
A mask removes the projected arm and pixels with invalid depth from the empty reference; in simulation the reference is rendered from the nominal camera with the movable objects removed.

\paragraph{Arm Anchor (metric).}
Forward kinematics at $\vect q_t$ places the sampled surface of every link in $B$, and registering this cloud to the back-projected camera depth yields $\hTbc$ metrically, with no learned component.
Two details make this work from a third-person view: \emph{visibility}, keeping only model points that project into the image and are not occluded by the observed depth, since the pedestal and the lower links are often hidden; and \emph{scoring}, ranking the trimmed-ICP~\cite{besl1992icp} solutions of a small hypothesis grid by a render-and-compare score that penalises pixels where the camera sees \emph{through} the arm model, which rejects degenerate solutions lying flat on the table.
Registering hidden model points of the arm to the table surface produces confident wrong answers, which is why only model points that project inside the image and are not occluded by the observed depth at the current hypothesis are registered.
Nearest-neighbour counts alone prefer degenerate solutions in which the arm model lies flat on the table; the render-and-compare score penalises, with double weight, pixels where the camera sees through the place where the arm should be, and rejects them.
The world anchor seeds the hypothesis grid of the arm anchor, so that a large camera shift does not move the arm far from its prior.

\paragraph{World Anchor (scene).}
Matching the live frame to the empty reference recovers $\dTc$ independently of the robot: SIFT~\cite{lowe2004sift} matches away from depth discontinuities are lifted to 3D and registered by RANSAC over closed-form fits~\cite{fischler1981ransac,umeyama1991least}, then refined by point-to-plane ICP~\cite{chen1992p2plane} on the static background.
The arm is masked and the objects are absent from the reference, so neither disturbs the estimate.

\paragraph{Fusion and Gate.}
With both anchors, \cref{eq:decomp} gives the base shift in closed form,
\begin{equation}
\hdTb^{(6)}=\Tbcz\,\hdTc\,\hTbc^{-1},\qquad \hdTb=\Pi_{\SE(2)}\big(\hdTb^{(6)}\big),
\label{eq:fuse}
\end{equation}
where $\Pi_{\SE(2)}$ keeps the in-plane translation and the yaw.
The discarded height, roll and pitch make the estimate checkable without ground truth: a base cannot leave the floor, so an out-of-plane residual above 2\,cm or $2^\circ$ means that an anchor has failed, and the episode then runs without canonicalisation; with fewer than eight world-anchor inliers only the image channel is disabled, and a noise floor of 5\,mm and $0.5^\circ$ protects the clean condition.
Both anchors span the out-of-plane residual, so the gate can detect but not isolate a failure: it cannot tell which anchor failed and therefore degrades globally, running the episode without canonicalisation.
For wide viewpoint changes over a bare table, an arm-only fallback registers the arm without a scene prior and accepts the fit only if the static reference depth agrees (\cref{sup:wide}).

\subsection{Channel-Separated Canonicalization}
\label{sec:method:canon}
\begin{figure*}[t]
\centering
\includegraphics[width=0.86\linewidth]{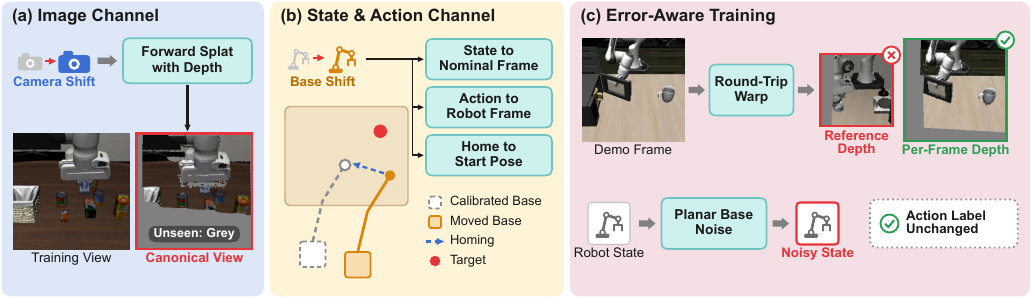}
\caption{\textbf{Canonicalization and error-aware training.} \textbf{(a)} Channel A forward-splats the shifted third-person image to the nominal camera with the measured depth; disocclusions and the out-of-frustum band are grey and are not hallucinated. \textbf{(b)} Channel B maps the reported end-effector pose into the nominal base frame with $\Cmat=\Twbz\hdTb(\Twbz)^{-1}$, rotates the policy's deltas back before execution, and homes the end effector to the start pose the demonstrations began from; in the nominal frame a base shift has become an object displacement on the table. \textbf{(c)} Training splats a demonstration frame to a random virtual camera and back with a perturbed shift drawn from the estimator's error model, which reproduces the holes, borders and residual misalignment of the test-time warp without changing the action label; the state receives a planar base-shift error about the base origin. Per-frame metric depth (right) removes the ghosting that a static reference depth leaves on the arm (middle).}
\label{fig:canon}
\end{figure*}

Given both shifts, the observation is mapped back to the demonstration distribution channel by channel (Proposition~3; illustrated in \cref{fig:canon}).

\paragraph{Channel A: Image.}
The third-person image is forward-splatted from $C$ to $C^0$ with $\hdTc$ alone: each pixel is back-projected with its measured depth, transformed, re-projected and rasterised with a z-buffer, so that the gripper and the objects move at their own depth.
Disocclusions and the band outside the original frustum are filled with grey, or with the empty reference for the wide views of LIBERO-Plus, and are never hallucinated; the wrist image is never warped.
When the estimated translation is below the noise floor, the warp reduces to the exact homography $\mathbf H=\mathbf K\,\Delta\mathbf R_c\,\mathbf K^{-1}$ and needs no depth.
A $10^\circ$ rotation leaves about a fifth of the frame unrecoverable at the border; we make no attempt to hallucinate it (\cref{fig:canon}a).

\paragraph{Channel B: State, Action, Start Pose.}
With $\Cmat=\Twbz\,\hdTb\,(\Twbz)^{-1}$, the reported end-effector pose is mapped into the nominal base frame by $\vect p\mapsto\mathbf R_{\Cmat}\vect p+\vect t_{\Cmat}$ and $\mathbf R\mapsto\mathbf R_{\Cmat}\mathbf R$, and the policy's deltas are rotated back before execution,
\begin{equation}
\Delta\vect p_b=\Delta\mathbf R_b^{\top}\Delta\vect p_{b^0},\qquad
\Delta\bm\omega_b=\Delta\mathbf R_b^{\top}\Delta\bm\omega_{b^0},
\label{eq:action}
\end{equation}
which is exact because $\mathbf R\Exp([\bm\omega]_\times)\mathbf R^{\top}=\Exp([\mathbf R\bm\omega]_\times)$.
In the nominal frame a base shift becomes an object displacement on the table, which the randomised demonstration layouts already cover, so no action label is changed.
Because the demonstrations start from one joint configuration, \method finally \emph{homes} the end effector to their start pose expressed in the current base frame, for at most 25 control steps and only when the estimated shift exceeds the estimator's own noise.
The quaternion of the canonicalised orientation is kept in the hemisphere of the input so that the downstream axis-angle does not flip near $\pi$, a real failure mode because the end effector points down; the result is normalised and tokenised exactly as in training.
Actions are rotated back after de-normalisation and before the controller's clipping, so the policy's normalisation statistics are untouched.
Homing drives the end effector with a proportional controller towards $\T{be}^{*}=\hdTb^{-1}\T{b^0e}(\vect q_0)$ for at most 25 control steps that count towards the episode budget, and only when the estimated shift exceeds $\max(3\sigma,\,1.5\,\mathrm{cm}/1.5^\circ)$, so that estimation noise never triggers it; it collects no data and uses no target.
Choosing $B^0$ rather than $C^0$ as the canonical frame is a convenience: the two differ by the constant $\Tbcz$, and any correct canonicalisation needs $\dTc$, which arm-only calibration cannot provide.

\subsection{Error-Aware Fine-Tuning}
\label{sec:method:train}
If $\hdTb=\dTb\Exp(\bm\epsilon)$, the canonicalised state is off by $\bm\epsilon_t+\bm\epsilon_\theta\times\vect p_e$ and the direction of a translational action by $\bm\epsilon_\theta$ only (Proposition~4); with $\|\vect p_e\|\approx0.6$\,m one degree of yaw costs about one centimetre, so rotation accuracy dominates.
A policy fine-tuned on pristine demonstrations has never seen a splatted image or a slightly-off state, and even ground-truth canonicalisation leaves a gap to the clean condition (\cref{sec:exp:main}).
\method closes it by reproducing the canonicaliser's output distribution on the demonstrations themselves (\cref{fig:overview}, bottom left).

\paragraph{Round-Trip Reprojection.}
With probability $p_{\mathrm{rt}}$ a training frame is splatted to a random virtual camera (up to 12\,cm and $12^\circ$ per axis) and back with a shift perturbed by the estimator's error model (8\,mm and $0.6^\circ$), using per-frame metric depth from Depth Anything V2~\cite{yang2024depthanything2} aligned to the calibration scale.
The result has exactly the holes, borders and residual misalignment of the test-time warp, and the action label is unchanged because a camera shift does not alter the correct base-frame action.
The per-frame Depth Anything V2 depth is aligned to the calibration scale on the static background of the reference frame; using the reference depth alone leaves ghosting on the arm (\cref{fig:canon}c, middle).

\paragraph{Structured State Noise.}
With probability $p_{\mathrm s}$ the state is transformed by a planar base-shift error $\Twbz\,\bm\delta_b\,(\Twbz)^{-1}$ (8\,mm and $0.6^\circ$), which couples position and yaw errors as Proposition~4 predicts; actions are left untouched, since with an unchanged image the direction of a state error is unobservable.
For the structured state noise, the axis-angle is chosen among its two equivalent representations so that it stays continuous.

\paragraph{Objective.}
The flow-matching objective of \pizf is unchanged and no token, head or parameter is added, so any \pizf checkpoint can run the \method front end without retraining; we fine-tune all weights with LeRobot~\cite{cadene2024lerobot}, a global batch of 32 and $p_{\mathrm{rt}}=p_{\mathrm s}=0.5$ (\cref{sup:details}).

\subsection{Front End for Wide Viewpoint Changes}
\label{sup:wide}
The LIBERO-Plus camera track moves the camera far beyond \bench: orbits of up to $75^\circ$ around the workspace, elevation changes, and distance changes of $2\times$, often over a bare table or a periodically textured floor.
The front end used there adds the following to the two anchors of \cref{sec:method:est}; on \bench it behaves like the basic version.

\paragraph{In-Scene Calibration and Scan.}
LIBERO-Plus ships its own scene assets, so the calibration bundle is recorded in its scenes: the nominal camera's empty-workspace RGB-D frame plus a short scan of twelve further views (azimuth $\pm20^\circ,\pm45^\circ,\pm70^\circ$ at the nominal height, $\pm45^\circ,\pm70^\circ$ raised by $15^\circ$, a view $1.6\times$ farther away, and one raised view), all looking at the same workspace point.
When the local solution is poor, the live frame is matched to every scan view (a wide baseline becomes a narrow one), and scoring and refinement use the fused static cloud of all views.
On a real robot this is a one-minute hand-held sweep of the empty table.

\paragraph{Plane-Constrained Global Search.}
Large orbits defeat feature matching, but the live and the reference view both contain the support plane.
Aligning it fixes tilt and height; the remaining yaw and in-plane translation are swept exhaustively (yaw in $2^\circ$ steps, translation by 2D FFT cross-correlation of height maps on a 2.5\,cm grid), keeping the best peak of every $10^\circ$ yaw bin, because symmetric layouts produce strong mirrored peaks.
Candidates are ranked by a dense score, the agreement of the live depth with the re-rendered static reference times the grey-value correlation on the agreeing points.

\paragraph{Floor Aliases and Plausibility.}
A periodically textured floor lets the scene lock on a solution one period away.
When the scene evidence is ambiguous and disagrees with the arm, the scene solution is translated along the floor plane and the translation with the best render-and-compare arm score is taken if it clearly beats the original, and only the image channel is enabled.
An implied base shift beyond 25\,cm or $30^\circ$, or an arm fit explaining fewer than 5\% of the model points, counts as inconsistent.

\paragraph{Image Channel.}
At these view changes a forward splat leaves cracks and large holes, so the image is warped backwards: the live depth is splatted into the nominal camera with a z-buffer, and every nominal pixel samples the live image where the two depths agree.
Pixels the live camera did not see are filled from the empty-scene reference rather than grey, which is also how the fine-tuning augmentation for LIBERO-Plus renders its round trips.

\paragraph{Arm-Only Fallback.}
If no estimate is trusted, the arm is registered without any scene prior: after plane alignment, yaw and in-plane translation are swept under the render-and-compare score, the best hypotheses are rescored on a local grid and polished by visibility-aware ICP, and the first hypothesis that explains enough of the arm and with which the static reference depth agrees is accepted.
It runs once per episode in a few seconds.

\subsection{Derivations}
\label{sup:derivations}

\paragraph{Decomposition (Proposition 1).}
With $\Twc=\Twcz\dTc$ and $\Twb=\Twbz\dTb$,
$\Tbc=\Twb^{-1}\Twc=\dTb^{-1}(\Twbz)^{-1}\Twcz\dTc=\dTb^{-1}\Tbcz\dTc$.
Solving for the base shift given the two anchors, $\dTb=\Tbcz\,\dTc\,\Tbc^{-1}$, which is \cref{eq:fuse}.

\paragraph{Action Conjugation (Proposition 3).}
The operational-space controller integrates a rotational delta as $\mathbf R\leftarrow\Exp([\Delta\bm\omega]_\times)\mathbf R$ in the frame in which the delta is expressed.
For a rotation $\mathbf R_\Delta$ between the nominal and the current base frame, $\mathbf R_\Delta\Exp([\bm\omega]_\times)\mathbf R_\Delta^{\top}=\Exp([\mathbf R_\Delta\bm\omega]_\times)$, so expressing a nominal-frame delta in the current frame is exactly $\Delta\bm\omega_b=\mathbf R_\Delta^{\top}\Delta\bm\omega_{b^0}$; no first-order approximation is involved.
The translational delta transforms as a free vector.

\paragraph{Error Propagation (Proposition 4).}
Write $\hdTb=\dTb\Exp(\bm\epsilon)$ and $\T{be}=(\mathbf R_e,\vect p_e)$.
To first order, $\hdTb\T{be}=\dTb\T{be}+\dTb\,[\bm\epsilon_t+\bm\epsilon_\theta\times\vect p_e;\ \bm\epsilon_\theta]$, so the canonicalised position is off by $\bm\epsilon_t+\bm\epsilon_\theta\times\vect p_e$ and the orientation by $\bm\epsilon_\theta$.
The de-canonicalised action $\mathbf R_\Delta^{\top}\Exp(-[\bm\epsilon_\theta]_\times)\Delta\vect p$ differs from the correct one by a rotation of $\bm\epsilon_\theta$ and does not depend on $\bm\epsilon_t$.

\section{Experiments}
\label{sec:experiments}

\subsection{Experimental Setup}
\label{sec:exp:setup}\label{sup:details}

\begin{table}[t]
\centering
\footnotesize
\setlength{\tabcolsep}{4pt}
\caption{\textbf{Perturbation Levels of \bench.} Camera translations are uniform in a ball and base translations in a disc, both between half and the full radius; rotations are uniform per axis. Last column: displacement of a target object in the $256^2$ third-person view for the largest translation / yaw.}
\label{tab:levels}
\begin{tabular}{lcccc c}
\toprule
 & \multicolumn{2}{c}{Camera $\dTc$} & \multicolumn{2}{c}{Base $\dTb$} & \\
\cmidrule(lr){2-3}\cmidrule(lr){4-5}
Level & trans. & rot./axis & trans. & yaw & px (t / yaw) \\
\midrule
L1 & $\le$2\,cm & $\pm3^\circ$ & $\le$2\,cm & $\pm3^\circ$ & 6 / 12 \\
L2 & $\le$5\,cm & $\pm5^\circ$ & $\le$5\,cm & $\pm8^\circ$ & 15 / 21 \\
L3 & $\le$10\,cm & $\pm10^\circ$ & $\le$10\,cm & $\pm15^\circ$ & 31 / 42 \\
\bottomrule
\end{tabular}
\end{table}

\begin{figure*}[t]
\centering
\includegraphics[width=0.98\linewidth]{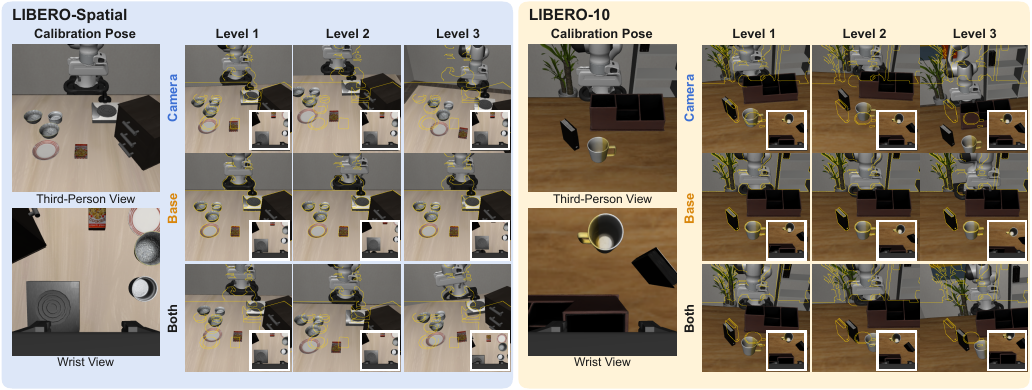}
\caption{\textbf{\bench Observations.} Episode 0 of a LIBERO-Spatial task (``pick up the black bowl between the plate and the ramekin and place it on the plate'') and of a LIBERO-10 task (``pick up the book and place it in the back compartment of the caddy''): third-person and wrist views at the calibration pose, and the third-person view with the wrist view inset under camera, base and compound shifts of level 1--3. Yellow: outline of the calibration view.}
\label{fig:bench_grid}
\end{figure*}

\begin{figure*}[t]
\centering
\includegraphics[width=0.98\linewidth]{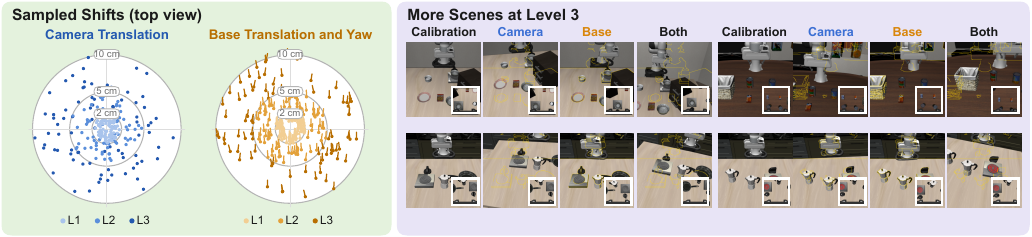}
\caption{\textbf{Shift Distribution and Further Scenes.} Left: 80 draws per level from the benchmark's sampler, seen from above: camera translations, and base translations with the sampled yaw drawn as a tick (rings at 2, 5 and 10\,cm). Right: four further tasks at the calibration pose and under level-3 shifts, drawn as in \cref{fig:bench_grid}.}
\label{fig:bench_design}
\end{figure*}

\begin{table}[t]
\centering
\footnotesize
\caption{\textbf{Feasibility of the Base Shifts}: share of sampled base poses with no robot--environment penetration after reset and a stable arm after settling (six samples per task and level). Invalid samples are kept in the evaluation, so they bound the success at base and compound L3.}
\label{tab:sup_feas}
\begin{tabular}{lccc}
\toprule
 & L1 & L2 & L3 \\
\midrule
all four suites (240 per level) & 1.00 & 1.00 & 0.97 \\
Spatial and LIBERO-10 (120 per level) & 1.00 & 1.00 & 0.94 \\
worst penetration (mm) & 0 & 0 & 30 \\
\bottomrule
\end{tabular}
\end{table}

\paragraph{Benchmarks.}
\textbf{\bench} extends LIBERO~\cite{liu2023libero}, whose tasks come with 50 demonstrations collected from one camera pose and one base pose, with a deterministic perturbation layer on its robosuite/MuJoCo environment~\cite{zhu2020robosuite,todorov2012mujoco}, so that all methods see identical scenes.
\emph{Camera} shifts rotate and translate the third-person camera; \emph{base} shifts move the robot in the floor plane and rotate it about the vertical, leaving table, objects and third-person camera fixed; \emph{both} applies the two independently.
Levels L1/L2/L3 bound the camera shift by 2/5/10\,cm and $3/5/10^\circ$ per axis and the base shift by 2/5/10\,cm and $3/8/15^\circ$ of yaw (\cref{tab:levels}); physically infeasible base poses are kept and bound the success at L3 (\cref{tab:sup_feas}).
Unless noted, state and actions follow the base-frame protocol of \cref{sec:method:pre}, which is bit-identical to the original data in the clean condition.
The main grid is LIBERO-Spatial and LIBERO-10 $\times$ 10 tasks $\times$ 10 conditions (clean plus three kinds at three levels) $\times$ 10 episodes, 2{,}000 episodes per model.
The \textbf{LIBERO-Plus camera track}~\cite{fei2026liberoplus} adds 1{,}599 variants with orbits of up to $75^\circ$ and distance changes of up to $2\times$, far outside our levels.
\cref{fig:bench_grid} shows what the policy observes in \bench for one task of each suite: the third-person and wrist views at the calibration pose, and the same episode under camera, base and compound shifts of the three levels of \cref{tab:levels}.
A camera shift displaces the whole third-person image and leaves the wrist view untouched; a base shift leaves the scene of the third-person image in place, moves the arm in it and changes the wrist view, and the state and the actions are expressed in the moved base frame.
\cref{fig:bench_design} shows the distribution the shifts are drawn from and four further tasks at level 3.

\paragraph{Shift Distribution and Feasibility.}
Because translations are drawn between half and the full radius of their level (\cref{tab:levels}), the translation ranges of the three levels do not overlap (\cref{fig:bench_design}, left): every L3 shift translates the camera or the base by at least 5\,cm, and at the largest L3 translation and yaw a target object moves by 31 and 42 pixels in the $256^2$ third-person view.
Large base shifts can make the robot penetrate the environment after reset.
\cref{tab:sup_feas} reports the share of sampled base poses that are physically valid: all of them at L1 and L2, and 97\% at L3 over all four suites (94\% on LIBERO-Spatial and LIBERO-10), with a worst penetration of 30\,mm.
These samples are kept rather than redrawn, so they bound the success at base and compound L3 for all methods alike.

\begin{figure*}[t]
\centering
\includegraphics[width=0.96\linewidth]{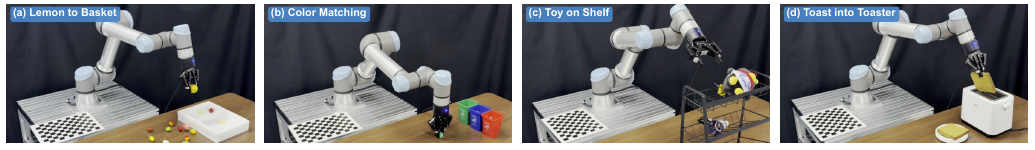}
\caption{\textbf{Real-Robot Tasks.} The four tabletop tasks form an increasing difficulty ladder: Lemon to Basket needs only coarse localisation; Color Matching adds an instruction-dependent referent and a cup opening only slightly wider than the block; Toy on Shelf requires carrying the toy over the shelf's wire frame and releasing it on the elevated upper tier; Toast into Toaster requires an insertion whose clearance is far below the object size, into a toaster that stays at the same spot across the demonstrations.}
\label{fig:tasks}
\end{figure*}

\paragraph{Real Robot.}
A UR5e with a Robotiq 2F-85 gripper on its own stand faces a table, observed by a RealSense D435i on a tripod (RGB-D) and a wrist D435i (RGB) (\cref{sup:hardware}).
Four tasks of increasing geometric tolerance, each with 50 teleoperated demonstrations at a single camera and base pose, are (a) \emph{Lemon to Basket}; (b) \emph{Color Matching}, placing the instructed block in the cup of the same colour; (c) \emph{Toy on Shelf}, placing a plush toy on the upper tier of a wire shelf; and (d) \emph{Toast into Toaster}, inserting a slice into a slot whose clearance is far below the object size (\cref{fig:tasks}).
A \emph{camera shift} moves the tripod; a \emph{base shift} moves the table with the objects and the tripod together by 8\,cm to the left and 6\,cm forward, away from the arm, which is exactly the base shift of \cref{eq:decomp} at the scale of L3; \emph{both} adds a tripod move.
The external camera is hand--eye calibrated once before data collection; re-calibration at the shifted poses serves only as ground truth for the oracle and the estimator's accuracy.
Every method runs 20 trials per task and condition on the same pre-generated layouts.

\paragraph{Policies and Baselines.}
All learned policies are fine-tuned from the public \pizf base checkpoint~\cite{pi2025pi05} with LeRobot~\cite{cadene2024lerobot} on the same demonstrations for the same number of steps.
\textbf{B1} is the official few-shot recipe; \textbf{W1} adds the augmentation of \cref{sec:method:train}; \method is W1 with the runtime.
\emph{B1+2D aug} adds the openpi crop-and-rotate augmentation; \emph{B2} trains on W1's samples but runs without canonicalisation; \emph{align-and-replay} runs our front end and channel B on B1, the closest instantiation of DINOBot~\cite{dipalo2024dinobot} with a closed-loop VLA; \emph{base gauge} and \emph{camera gauge} use the arm anchor only (\cref{sec:method:pre}); \emph{GT oracle} canonicalises with the true shifts and bounds any estimator.
\emph{Avg} is the mean over the shifted conditions; differences below two points lie within the Wilson 95\% intervals~\cite{wilson1927probable} and are not interpreted.

\begin{table}[t]
\centering
\footnotesize
\caption{\textbf{Runtime} of the front end (median over 36 resets, one CPU thread) and of the canonicaliser, and homing statistics (12 episodes per condition). Policy inference was timed on a GPU shared with a training run and is an upper bound.}
\label{tab:sup_runtime}
\resizebox{\linewidth}{!}{%
\begin{tabular}{lc}
\toprule
Component & Time \\
\midrule
World anchor (SIFT + RANSAC) & 124\,ms \\
World anchor refinement (point-to-plane ICP) & 83\,ms \\
Arm anchor (multi-hypothesis ICP) & 524\,ms \\
Whole front end, once per episode & 771\,ms \\
Channel A warp (per control step, when applied) & 11\,ms \\
\pizf inference (per chunk of 10 steps) & $\leq$320\,ms \\
\midrule
Homing triggered: base L2 / compound L3 / clean & 12/12 / 8/12 / 2/12 \\
Homing steps when triggered: mean / max & 6.9 / 9 (base), 11.8 / 19 (compound) \\
\bottomrule
\end{tabular}}
\end{table}

\paragraph{Model Training.}
All weights of \pizf are fine-tuned with LeRobot: vocabulary embedding frozen, bf16 parameters, gradient checkpointing, global batch 32, AdamW with learning rate $2.5{\times}10^{-5}$, 1k warm-up steps and cosine decay to $2.5{\times}10^{-6}$, $n_{\text{action\_steps}}=10$, one empty camera slot, and the discretised state written into the prompt as in the official recipe.
B1, B2 and W1 start from the public base checkpoint and train for 10k steps on four A800 GPUs with DDP, a dtype-split ZeRO-1 optimiser state and micro-batch accumulation; the few-shot runs (\cref{tab:sup_fewshot}) use the first $K$ demonstrations of every task with the same recipe.
For LIBERO-Plus the public LIBERO checkpoint is fine-tuned further for 10k steps (300 warm-up steps, global batch 32, learning rate $2.5{\times}10^{-5}$ with cosine decay), with the augmentation rendered like the runtime's image channel (inverse warp, unseen pixels from the reference).
The augmentation is applied inside the dataset with $p_{\mathrm{rt}}=p_{\mathrm s}=0.5$.
Real-robot policies are single multi-task policies fine-tuned from the same base checkpoint on the four tasks jointly for 8\,k steps; the round-trip augmentation of W1 uses the D435i depth, with invalid pixels filled from their nearest valid neighbours, instead of predicted depth.

\paragraph{Policy Evaluation.}
Simulation: 10 episodes per task and condition on LIBERO-Spatial and LIBERO-10 (200 per cell), one perturbation seed shared by all methods; LIBERO-Plus: one episode per variant, three evaluation seeds.
Real robot: 20 trials per task and condition at the condition's pose; a trial counts as successful only when the complete task-specific criterion is satisfied without human intervention, within 60\,s.
Trials are run in blocks per pose: once a pose is set up, all methods are evaluated on the same pre-generated layouts in an interleaved order, so that slow drifts in lighting or placement affect every method equally.
All compared policies share workspace bounds, velocity limits and emergency-stop conditions, so that differences reflect the policy rather than the execution stack.

\paragraph{Runtime.}
Estimation runs once at reset on the CPU and again after homing; channel A costs one forward splat per control step (\cref{tab:sup_runtime}).
The estimator, the canonicaliser and the augmentation amount to about 2k lines of NumPy on top of a LeRobot environment plugin.

\paragraph{Instruction Protocol.}
The released LIBERO-Plus evaluation builds each policy prompt from the variant's file name, which appends the variant's parameters to the instruction, as also noted by~\cite{yan2026flexpi}.
We report both this \emph{released} protocol and a \emph{corrected} one that passes the task's own instruction; the suffix is stripped by a regular expression that maps all 1{,}599 camera tasks back to an original LIBERO instruction.

\subsection{Main Results}
\label{sec:exp:main}
We ask three questions: (1) How much of the loss caused by a pose shift can canonicalisation recover in principle, and how much does \method recover with an estimated shift? (2) Is the world anchor necessary? (3) Does the advantage transfer to an external benchmark, to scarcer demonstrations and to a real robot?

\begin{table}[t]
\centering
\footnotesize
\setlength{\tabcolsep}{3pt}
\caption{\textbf{Sensitivity and Oracle Bound.} The public \pizf checkpoint on LIBERO-Spatial and LIBERO-10 (200 episodes per cell; cells are L1 / L2 / L3; base-frame protocol unless noted). The \method runtime is applied without retraining. $^*$World-frame average over base and compound rows only, since a camera shift is protocol-independent.}
\label{tab:oracle}
\resizebox{\linewidth}{!}{%
\begin{tabular}{lcccc}
\toprule
Kind & None & \makecell{World-frame\\protocol} & \makecell{\method\\runtime} & \makecell{GT\\oracle} \\
\midrule
clean & 96.0 & -- & 97.5 & -- \\
camera & 96.0\,/\,97.0\,/\,82.5 & -- & 97.5\,/\,94.5\,/\,86.0 & 96.5\,/\,97.0\,/\,92.5 \\
base & 97.0\,/\,92.0\,/\,74.0 & 94.0\,/\,97.0\,/\,85.5 & 95.5\,/\,94.5\,/\,91.0 & 95.5\,/\,94.0\,/\,91.0 \\
both & 95.5\,/\,89.0\,/\,60.5 & 94.5\,/\,92.0\,/\,73.0 & 95.5\,/\,90.5\,/\,80.5 & 96.5\,/\,93.5\,/\,89.5 \\
\midrule
\rowcolor{meanblue}
Avg over shifts & 87.1 & 89.3$^*$ & \textbf{91.7}\,\gain{+4.6} & 94.0 \\
\bottomrule
\end{tabular}}
\end{table}

\paragraph{Sensitivity and Oracle Bound.}
\cref{tab:oracle} evaluates the public \pizf LIBERO checkpoint, the strongest available policy.
Thanks to its crop-and-rotate augmentation and the wrist camera, L1 and L2 camera shifts cost it nothing, but success falls to 82.5\% at camera L3 and to 74.0\% and 60.5\% at base and compound L3: base shifts, which no prior benchmark measures, are the harder half of the problem even for the best policy.
With the true shifts the oracle returns every condition to within 6.5 points of clean, and the \method runtime alone, with no retraining and no ground truth, recovers 67\% of the oracle gain on average (87.1 $\rightarrow$ 91.7 of 94.0), matches the oracle at base L3 and adds 20.0 points at compound L3, while leaving the clean condition intact (97.5 vs.\ 96.0).
The world-frame column quantifies the protocol leakage: base L3 costs 22.0 points under the base-frame protocol but only 10.5 in LIBERO's world frame, so base-robustness numbers measured in the native frame should be read with care.

\begin{table*}[t]
\centering
\footnotesize
\setlength{\tabcolsep}{4.2pt}
\renewcommand{\arraystretch}{1.12}
\caption{\textbf{Pose-Shift Robustness on \bench} (LIBERO-Spatial and LIBERO-10, 200 episodes per cell, base-frame protocol). All learned rows share checkpoint, demonstrations and budget; \emph{est.}: shift estimated at test time; $\dagger$: upper bound with ground-truth poses. Red: gain over B1. Base / camera gauge: the arm's pose change read as a camera / base shift.}
\label{tab:main}
\begin{tabular}{l cc c ccc ccc ccc c}
\toprule
 & \multicolumn{2}{c}{Requires} & & \multicolumn{3}{c}{Camera} & \multicolumn{3}{c}{Base} & \multicolumn{3}{c}{Both} & \\
\cmidrule(lr){2-3}\cmidrule(lr){5-7}\cmidrule(lr){8-10}\cmidrule(lr){11-13}
Method & \makecell{extra\\data} & \makecell{run-\\time} & Clean & L1 & L2 & L3 & L1 & L2 & L3 & L1 & L2 & L3 & Avg \\
\midrule
B1 (official few-shot recipe) & -- & -- & 84.5 & 82.0 & 83.5 & 67.5 & 84.5 & 80.0 & 63.0 & 87.5 & 73.5 & 44.5 & 74.0 \\
B1 + 2D aug (crop, rotate) & -- & -- & 86.5 & 87.0 & 83.0 & 75.0 & 87.5 & 82.5 & 66.5 & 84.0 & 79.0 & 53.0 & 77.5\,\gain{+3.5} \\
B2 (W1 data, no runtime) & -- & -- & 88.5 & 87.5 & 86.5 & 80.0 & 88.0 & 82.0 & 72.0 & 88.0 & 80.5 & 65.0 & 81.1\,\gain{+7.1} \\
\midrule
Align-and-replay (B1 + front end, ch.\ B) & -- & est. & 86.0 & 82.5 & 80.0 & 65.5 & 86.5 & 90.0 & 83.0 & 86.0 & 77.0 & 63.0 & 79.3\,\gain{+5.3} \\
Base gauge (B1 + arm anchor only) & -- & est. & 86.5 & 85.0 & 80.5 & 64.0 & 87.5 & 74.0 & 56.0 & 82.5 & 73.0 & 46.0 & 72.1\,\gain{$-$1.9} \\
Camera gauge (B1 + arm anchor only) & -- & est. & 91.5 & 83.5 & 81.5 & 64.0 & 89.0 & 88.5 & 84.0 & 88.0 & 72.5 & 49.0 & 77.8\,\gain{+3.8} \\
\method runtime on B1 & -- & est. & 88.0 & 88.5 & 83.5 & 73.5 & 84.5 & 87.0 & 86.5 & 86.5 & 77.0 & 66.0 & 81.4\,\gain{+7.4} \\
\method runtime on B1 + 2D aug & -- & est. & 86.0 & 88.0 & 82.5 & 81.5 & 90.5 & 87.0 & 87.0 & 88.5 & 83.5 & 69.0 & 84.2\,\gain{+10.2} \\
\rowcolor{oursblue}
\textbf{\method} (W1 + runtime) & -- & est. & \textbf{87.0} & \textbf{91.0} & \textbf{87.0} & \textbf{84.5} & \textbf{89.5} & \textbf{89.5} & \textbf{86.5} & \textbf{91.0} & \textbf{86.5} & \textbf{79.0} & \textbf{87.2}\,\gain{+13.2} \\
\midrule
GT oracle on B1$^\dagger$ & -- & GT & 86.0 & 85.0 & 86.0 & 75.0 & 89.5 & 85.5 & 82.0 & 84.5 & 84.5 & 75.0 & 83.0 \\
GT oracle on W1$^\dagger$ & -- & GT & 89.0 & 89.5 & 88.5 & 86.5 & 90.0 & 91.0 & 86.5 & 91.0 & 87.5 & 87.5 & 88.7 \\
\bottomrule
\end{tabular}
\end{table*}

\paragraph{Pose-Shift Robustness.}
\cref{tab:main} presents the main comparison.
\textbf{The standard recipe is fragile:} B1 loses 17.0 points at camera L3, 21.5 at base L3 and 40.0 at compound L3, and the 2D augmentation recovers part of the camera axis only.
\textbf{Data alone is not enough:} B2 sees exactly W1's samples and gains 7.1 points over B1, but stays 6.1 behind \method, because the policy cannot undo a shift it never observes explicitly.
\textbf{Estimation plus error-aware training closes the loop:} the runtime on B1 already recovers 82\% of the gap to B1's own oracle, and with error-aware training \method reaches 87.2\%, 13.2 points above B1 and within 1.5 points of its own oracle, with a clean score not below B1's.
\method has the best average of all methods that receive no ground truth, and its margin over B1 grows with the level: 5.8, 8.7 and 25.0 points at L1, L2 and L3.

\paragraph{Which Anchor Matters.}
The single-anchor gauges fail as the gauge ambiguity predicts: the base gauge under base shifts (base L3: 56.0 vs.\ 86.5 for \method), and both gauges at camera L3 (64.0), because without the world anchor's seed the arm registration itself misses large camera shifts; only the two-anchor estimate handles the compound condition (both L3: 79.0 vs.\ 46.0 and 49.0).
Align-and-replay recovers the base axis (base L3: 83.0) but not the camera axis (camera L3: 65.5), so both channels are needed.

\begin{table}[t]
\centering
\footnotesize
\setlength{\tabcolsep}{4pt}
\caption{\textbf{LIBERO-Plus Camera Track} (1{,}599 tasks, one episode each). \emph{Released}: the benchmark's prompts, which append the variant's parameters; \emph{corrected}: the task's own instruction~\cite{yan2026flexpi}. $^\ddagger$Prompt handling not stated. Ours: evaluation seed 0 / mean over three evaluation seeds.}
\label{tab:liberoplus}
\resizebox{\linewidth}{!}{%
\begin{tabular}{lccc}
\toprule
Method & Perturbed data & Released & Corrected \\
\midrule
OpenVLA-OFT~\cite{fei2026liberoplus} & -- & 56.4 & -- \\
\pizero-FAST~\cite{fei2026liberoplus} & -- & 65.1 & -- \\
CycleVLA (\pizf)~\cite{ma2026cyclevla} & -- & 74.2$^\ddagger$ & -- \\
Cross-View AC~\cite{huang2026crossview} & sim re-render & 87.2$^\ddagger$ & -- \\
TBD-VLA~\cite{lee2026tbdvla} & -- & 87.8$^\ddagger$ & -- \\
Flex-$\pi$~\cite{yan2026flexpi} & -- & -- & 91.6 \\
OpenVLA-OFT+~\cite{fei2026liberoplus} & 20k traj. & 92.8 & -- \\
\midrule
\pizf public checkpoint & -- & 62.5 & 63.8 \\
\quad + \method runtime (no training) & -- & 89.4\,/\,88.0 & 88.1\,/\,87.7 \\
\rowcolor{oursblue}
\textbf{\method} (error-aware fine-tuning + runtime) & -- & \textbf{93.7}\,/\,\textbf{92.4} & \textbf{93.7}\,/\,\textbf{92.0} \\
\bottomrule
\end{tabular}}
\end{table}

\paragraph{Cross-Benchmark Transfer.}
On the LIBERO-Plus camera track (\cref{tab:liberoplus}) the public checkpoint scores 62.5\%.
The \method runtime alone, with a calibration bundle and a short scan of the deployment scene (\cref{sup:wide}), lifts it to 89.4\% without retraining or extrinsics, and error-aware fine-tuning brings \method to 93.7\% (92.4\% over three seeds), the best released-protocol result among methods without perturbed training data, ahead of Cross-View AC (87.2\%) and TBD-VLA (87.8\%); only OpenVLA-OFT+ (92.8\%), trained on 20k trajectories from the benchmark's own perturbation distribution, remains comparable.
With the task's own instruction instead of the released prompt, which leaks the variant's parameters~\cite{yan2026flexpi}, \method scores 93.7\% (92.0\%) against 91.6\% for Flex-$\pi$.
The estimate is accurate and applied in 89\% of the tasks, which then succeed 94.3\% of the time; most remaining failures lie at the widest orbits, where over half of the frame is outside the original frustum (\cref{tab:lpfail}).

\paragraph{Few-Shot Scaling.}
The advantage of \method grows as demonstrations become scarcer, from 13.2 points over B1 with 50 demonstrations per task to 15.7 with 20 and 19.2 with 10, so that \method with 10 demonstrations (82.4\%) exceeds the standard recipe with 50 (74.0\%); the runtime and the augmentation both contribute at every data size (\cref{fig:fewshot}, \cref{tab:sup_fewshot}).

\begin{table}[t]
\centering
\footnotesize
\caption{\textbf{Few-Shot Scaling}: average shifted success (nine conditions) on LIBERO-Spatial and LIBERO-10 with the first 10, 20 or all 50 demonstrations per task, same recipe and budget.}
\label{tab:sup_fewshot}
\begin{tabular}{lccc}
\toprule
Demonstrations per task & 10 & 20 & 50 \\
\midrule
B1 (standard recipe) & 63.2 & 66.6 & 74.0 \\
B1 + \method runtime & 73.3 & 76.3 & 81.4 \\
W1 (error-aware training only) & 71.7 & 73.3 & 81.1 \\
\rowcolor{oursblue}
\textbf{\method} (W1 + runtime) & \textbf{82.4} & \textbf{82.3} & \textbf{87.2} \\
\midrule
gain of \method over B1 & \gain{+19.2} & \gain{+15.7} & \gain{+13.2} \\
\bottomrule
\end{tabular}
\end{table}

\begin{table}[t]
\centering
\footnotesize
\setlength{\tabcolsep}{4pt}
\caption{\textbf{Real-Robot Results} (UR5e, success rate in \%, 20 trials per task and condition, averaged over the four tasks; per-task numbers in \cref{tab:sup_real}). Avg is the mean over the three shifted conditions; red superscripts are gains over B1; $\dagger$ uses the re-calibrated ground-truth shifts and is an upper bound.}
\label{tab:real}
\resizebox{\linewidth}{!}{%
\begin{tabular}{lccccc}
\toprule
Method & Nominal & Camera & Base & Both & Avg \\
\midrule
\pizf (B1) & 92.5 & 60.0 & 40.0 & 30.0 & 43.3 \\
B1 + 2D aug & 93.8 & 72.5 & 42.5 & 32.5 & 49.2\,\gain{+5.9} \\
\method runtime on B1 & 91.3 & 75.0 & 72.5 & 62.5 & 70.0\,\gain{+26.7} \\
\rowcolor{oursblue}
\textbf{\method} (W1 + runtime) & \textbf{95.0} & \textbf{85.0} & \textbf{80.0} & \textbf{75.0} & \textbf{80.0}\,\gain{+36.7} \\
\midrule
GT oracle on W1$^\dagger$ & 95.0 & 87.5 & 86.3 & 82.5 & 85.4 \\
\bottomrule
\end{tabular}}
\end{table}

\begin{figure}[t]
\centering
\includegraphics[width=\linewidth]{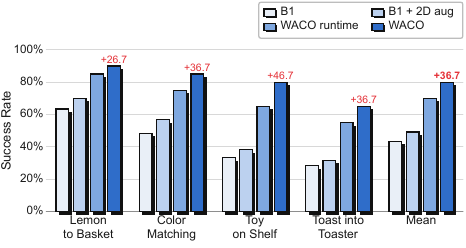}
\caption{\textbf{Real-Robot Per-Task Performance.} Success under pose shifts (mean of the camera, base and compound conditions) on the four UR5e tasks. \method improves every task, most on Toy on Shelf (+46.7 points).}
\label{fig:real}
\end{figure}

\paragraph{Real-Robot Deployment.}
\cref{tab:real} and \cref{fig:real} report the UR5e results.
In the nominal condition all methods are within noise of each other, so the gate leaves a correctly placed camera untouched.
Under shifts B1 degrades with the required tolerance, from 63\% on Lemon to Basket to 28\% on Toast into Toaster, and fails most under base and compound shifts, where its proprioception lives in a frame that has moved; on Toast into Toaster it reaches for the slot's old position in almost every base-shift trial, and the 2D augmentation helps only under camera shifts.
\method raises the average shifted success from 43\% to 80\%, improves every task under every condition, most on Toy on Shelf (+46.7 points), and comes within 5.4 points of the ground-truth oracle.
The runtime alone recovers 26.7 points and error-aware training adds 10 more, mirroring the decomposition on \bench.
The front end runs once per trial in about 0.8\,s on the robot's CPU and recovers the real shifts with median errors of 2.8\,mm and $0.3^\circ$ for the camera and 3.8\,mm and $0.4^\circ$ for the base, comparable to the repeatability of hand--eye re-calibration itself (\cref{tab:real_est}).

\begin{table}[t]
\centering
\footnotesize
\setlength{\tabcolsep}{4pt}
\caption{\textbf{Real-Robot Analysis} on Toy on Shelf and Toast into Toaster (20 trials per task and condition, averaged over the two tasks). \emph{Gauges}: arm anchor only, the change read as a camera (base gauge) or base shift (camera gauge). \emph{Few-shot}: the first 10 of 50 demonstrations per task. \emph{Occluded}: a box hides about a third of the static background.}
\label{tab:real_analysis}
\begin{tabular}{llcccc}
\toprule
Block & Method & Camera & Base & Both & Avg \\
\midrule
\multirow{3}{*}{Gauges} & Base gauge & 55 & 25 & 20 & 33.3 \\
 & Camera gauge & 30 & 65 & 35 & 43.3 \\
 & \cellcolor{oursblue}\textbf{\method} & \cellcolor{oursblue}\textbf{77.5} & \cellcolor{oursblue}\textbf{72.5} & \cellcolor{oursblue}\textbf{67.5} & \cellcolor{oursblue}\textbf{72.5} \\
\midrule
\multirow{2}{*}{\makecell[l]{Few-shot\\(10 demos)}} & B1 & 45 & 25 & 15 & 28.3 \\
 & \cellcolor{oursblue}\textbf{\method} & \cellcolor{oursblue}\textbf{72.5} & \cellcolor{oursblue}\textbf{67.5} & \cellcolor{oursblue}\textbf{62.5} & \cellcolor{oursblue}\textbf{67.5}\,\gain{+39.2} \\
\midrule
\multirow{2}{*}{Occluded} & \method w/o gate & -- & 40 & -- & -- \\
 & \cellcolor{oursblue}\textbf{\method} & \cellcolor{oursblue}-- & \cellcolor{oursblue}\textbf{72.5} & \cellcolor{oursblue}-- & \cellcolor{oursblue}-- \\
\bottomrule
\end{tabular}
\end{table}

\paragraph{Real-Robot Analysis.}
\cref{tab:real_analysis} repeats the key findings on the two tasks with the tightest tolerance.
The base gauge fails under base shifts (25 vs.\ 72.5 for \method) and the camera gauge under camera shifts (30 vs.\ 77.5), while the two-anchor estimate handles all three conditions.
With only 10 demonstrations per task, B1 falls to 28.3\% under shifts while \method keeps 67.5\%, a gain of 39.2 points that is almost as large as with 50 demonstrations (41.7).
When a box occludes about a third of the static background, the gate disables the image channel and the base channel falls back on the arm anchor, which is correct because the camera has not moved relative to the table: \method keeps 72.5\%, whereas applying the occluded estimates drops it to 40\%.

\subsection{Ablation Results}
\label{sec:exp:ablation}

\begin{table}[t]
\centering
\footnotesize
\setlength{\tabcolsep}{3pt}
\renewcommand{\arraystretch}{1.10}
\caption{\textbf{Ablations.} Each block changes one component of the full method (blue row) on LIBERO-Spatial and LIBERO-10. The full row is the same evaluation as \method in \cref{tab:main}. Avg is the mean over the nine shifted conditions.}
\label{tab:ablation}
\resizebox{\linewidth}{!}{%
\begin{tabular}{llcccc}
\toprule
Block & Variant & Clean & Cam L3 & Base L3 & Avg \\
\midrule
\rowcolor{oursblue}
-- & \textbf{\method} (full) & \textbf{87.0} & \textbf{84.5} & \textbf{86.5} & \textbf{87.2} \\
\midrule
\multirow{3}{*}{Channels} & A only (image) & 87.0 & 82.5 & 67.5 & 82.0 \\
 & B only (state, action) & 87.0 & 77.5 & 81.5 & 83.6 \\
 & B without homing & 87.0 & 77.5 & 78.5 & 81.7 \\
\midrule
\multirow{3}{*}{Anchors} & arm only ($\dTc{=}I$) & 87.0 & 78.5 & 86.5 & 82.6 \\
 & scene only ($\dTb{=}I$) & 87.0 & 84.0 & 66.5 & 81.7 \\
 & no consistency gate & 87.0 & 77.5 & 84.0 & 84.5 \\
\midrule
\multirow{3}{*}{Training} & runtime on B1 & 88.0 & 73.5 & 86.5 & 81.4 \\
 & round-trip only & 86.5 & 80.0 & 86.5 & 86.4 \\
 & state noise only & 88.0 & 68.0 & 83.5 & 80.3 \\
\midrule
\multirow{2}{*}{Depth} & reference depth only & 87.0 & 78.0 & 83.5 & 85.9 \\
 & plane homography only & 87.0 & 84.0 & 86.5 & 86.4 \\
\bottomrule
\end{tabular}}
\end{table}

\paragraph{Component Ablation.}
\cref{tab:ablation} removes one component at a time.
\emph{Channels:} the image channel alone leaves the base axis blind (base L3: 67.5) and the state channel alone the camera axis (camera L3: 77.5); only both handle the compound condition, as the channel separation predicts, and homing adds 1.9 points on average.
\emph{Anchors:} the arm anchor alone fails under compound shifts (both L3: 66.0), the scene anchor alone leaves the base channel blind (base L3: 66.5), and without the gate the wrong estimates it would have vetoed cost 2.7 points.
\emph{Training:} round-trip reprojection is the main component (86.4 on its own). State noise alone averages 80.3, with camera L3 at 68.0, next to B1, and base L3 at 83.5. The combination is the strongest variant (87.2).
\emph{Depth:} the static reference depth or a plane homography instead of the measured depth costs only 1.3 and 0.8 points, so at these shifts the warp is not the bottleneck.

\begin{figure}[t]
\centering
\includegraphics[width=0.98\linewidth]{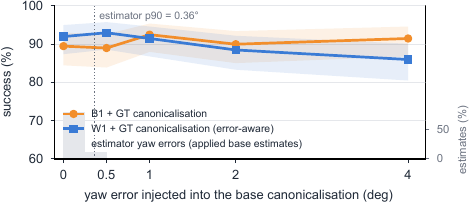}
\caption{\textbf{Sensitivity to Calibration Error} (base L3, LIBERO-Spatial, 200 episodes per point, 95\% Wilson bands): yaw error injected into the ground-truth base canonicalisation, and the histogram of the yaw errors of the estimator's applied base estimates.}
\label{fig:errcurve}
\end{figure}

\paragraph{Estimator Accuracy and Calibration Error.}
Over 240 simulated resets (\cref{tab:estimator}) the world anchor recovers the camera shift with a median error of 0.3\,mm and $0.0^\circ$, and the fused base shift has a median error of 2.4\,mm and $0.1^\circ$, inside the 7\,mm and $1^\circ$ that Proposition~4 sets for sub-centimetre state error.
Injecting yaw error into the true base shift (\cref{fig:errcurve}), neither policy collapses up to $4^\circ$, whereas the 90th percentile of the applied base-yaw error is $0.36^\circ$ in simulation and the robot's base-yaw error reaches $1.2^\circ$ at the 90th percentile: estimation accuracy is not the limiting factor, and the gain of error-aware training comes from the image artefacts of canonicalisation, as the oracle on W1 exceeds the oracle on B1.

\begin{table}[t]
\centering
\footnotesize
\setlength{\tabcolsep}{3pt}
\caption{\textbf{Estimator Accuracy and Gate Behaviour} over 240 resets (four suites, three tasks each, two episodes per condition): median / 90th-percentile error against the true shift. A catastrophic estimate exceeds 2\,cm or $1.5^\circ$; \emph{caught} counts catastrophic estimates whose channel the gate disabled, \emph{false} counts accurate estimates of a real shift that it disabled.}
\label{tab:estimator}
\resizebox{\linewidth}{!}{%
\begin{tabular}{lcccccc}
\toprule
 & \multicolumn{2}{c}{$\hdTc$ error} & \multicolumn{2}{c}{$\hdTb$ error} & \multicolumn{2}{c}{Gate} \\
\cmidrule(lr){2-3}\cmidrule(lr){4-5}\cmidrule(lr){6-7}
Condition & mm & deg & mm & deg & caught & false \\
\midrule
clean         & 0.0\,/\,6.6  & 0.0\,/\,0.0 & 0.7\,/\,7.6  & 0.1\,/\,0.2 & 0/0   & 0 \\
camera L1--L3 & 0.8\,/\,14.5 & 0.0\,/\,0.3 & 1.7\,/\,18.5 & 0.2\,/\,1.0 & 3/13  & 1 \\
base L1--L3   & 0.0\,/\,3.9  & 0.0\,/\,0.0 & 2.9\,/\,10.6 & 0.1\,/\,0.3 & 4/4   & 0 \\
both L1--L3   & 0.8\,/\,17.2 & 0.0\,/\,1.0 & 4.8\,/\,25.0 & 0.2\,/\,2.5 & 16/24 & 8 \\
\midrule
\rowcolor{meanblue}
all           & 0.3\,/\,10.7 & 0.0\,/\,0.2 & 2.4\,/\,19.0 & 0.1\,/\,1.2 & 23/41 & 9 \\
\bottomrule
\end{tabular}}
\end{table}

\begin{table}[t]
\centering
\footnotesize
\setlength{\tabcolsep}{3pt}
\caption{\textbf{Real-Robot Estimator Accuracy} against hand--eye re-calibration (median / 90th-percentile error; 10 poses per condition, three arm configurations each, 120 estimates). \emph{Accepted}: share of estimates the gate applies.}
\label{tab:real_est}
\resizebox{\linewidth}{!}{%
\begin{tabular}{lccccc}
\toprule
 & \multicolumn{2}{c}{$\hdTc$ error} & \multicolumn{2}{c}{$\hdTb$ error} & \\
\cmidrule(lr){2-3}\cmidrule(lr){4-5}
Condition & mm & deg & mm & deg & Accepted \\
\midrule
nominal & 0.6\,/\,2.4 & 0.0\,/\,0.1 & 1.1\,/\,3.8 & 0.1\,/\,0.2 & 100\% \\
camera shift & 3.4\,/\,11.8 & 0.4\,/\,1.2 & 4.2\,/\,14.5 & 0.5\,/\,1.6 & 86.7\% \\
base shift & 1.9\,/\,5.7 & 0.2\,/\,0.4 & 3.6\,/\,10.2 & 0.3\,/\,0.9 & 96.7\% \\
both & 4.7\,/\,16.5 & 0.7\,/\,1.8 & 5.5\,/\,19.8 & 0.8\,/\,2.1 & 83.3\% \\
\midrule
\rowcolor{meanblue}
all & 2.8\,/\,11.2 & 0.3\,/\,0.9 & 3.8\,/\,13.7 & 0.4\,/\,1.2 & 91.7\% \\
\bottomrule
\end{tabular}}
\end{table}

\paragraph{Estimator Accuracy in Simulation.}
After point-to-plane refinement the world anchor's rotation error is essentially zero, and the tail of its translation error (10.7\,mm at the 90th percentile) comes from sliding along a bare table plane, which the flat background constrains weakly.
The tail of the fused base shift is set by the compound condition (25.0\,mm and $2.5^\circ$ at the 90th percentile), where a grazing view of the arm is fused with a large camera shift.
The arm anchor is the accurate half of the pair because the arm is a rigid, well-textured, metrically known object.
The gate disables the channel of 23 of the 41 catastrophic estimates and wrongly disables 9 accurate ones, which merely run the episode uncanonicalised; ten of the 18 catastrophic estimates that pass the gate are 11--16\,cm slides along a bare table with under $2^\circ$ of rotation error.
Without the gate, wrong estimates are applied in 23 more of the 240 resets, which a deployed system must avoid.

\paragraph{Sensitivity to Calibration Error.}
In \cref{fig:errcurve}, B1 stays between 89 and 93\% up to $4^\circ$ of injected yaw error and W1 loses 6 points at $4^\circ$, because the closed-loop policy corrects small frame errors from the images.

\paragraph{Estimator Accuracy on the Real Robot.}
Measured against hand--eye re-calibration at 10 poses per condition and three arm configurations (120 estimates; \cref{tab:real_est}), the camera shift is recovered with a median error of 2.8\,mm and $0.3^\circ$ and the base shift with 3.8\,mm and $0.4^\circ$, comparable to the repeatability of the re-calibration itself (1.8\,mm, $0.3^\circ$); real depth noise widens the tails but leaves the medians inside the flat region of \cref{fig:errcurve}, and the gate accepts 91.7\% of the estimates.
The empty-table reference captured at calibration sufficed for the world anchor despite illumination changes over the evaluation days.

\section{Conclusion}
\label{sec:conclusion}
We introduced \method, a world-anchored canonicalization framework that makes few-shot vision-language-action policies robust to camera and robot-base shifts without test-time calibration, perturbed training data or architectural change.
Treating the arm as a metric calibration target and the static scene as the world anchor, \method recovers both shifts from a single RGB-D frame, canonicalises image and proprioception separately, and trains the policy for the estimator it runs with.
In simulation and on a real UR5e it outperforms the standard few-shot recipe and prior viewpoint-robust methods, most where demonstrations are scarcest.

\bibliographystyle{IEEEtran}
\bibliography{main}

\twocolumn[{%
\begin{center}
{\LARGE Towards Pose-Robust Few-Shot Vision-Language-Action Models}\\[6pt]
{\Large Supplementary Material}\\[14pt]
\end{center}}]
\setcounter{section}{0}
\renewcommand{\thesection}{S\arabic{section}}
\renewcommand{\theHsection}{S\arabic{section}}
\setcounter{table}{0}
\renewcommand{\thetable}{S\arabic{table}}
\renewcommand{\theHtable}{S\arabic{table}}
\setcounter{figure}{0}
\renewcommand{\thefigure}{S\arabic{figure}}
\renewcommand{\theHfigure}{S\arabic{figure}}
\setcounter{equation}{0}
\renewcommand{\theequation}{S\arabic{equation}}
\renewcommand{\theHequation}{S\arabic{equation}}

\section{Hardware Setup}
\label{sup:hardware}

\begin{figure}[t]
\centering
\includegraphics[width=\linewidth]{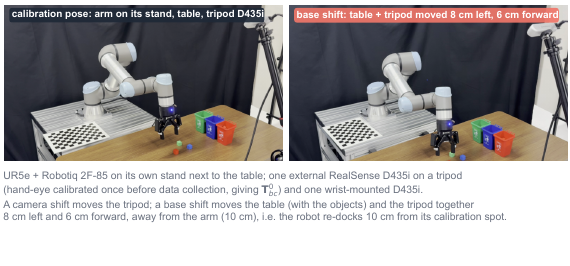}
\caption{\textbf{UR5e Setup.} The workspace at the calibration pose (left) and after the base shift (right); the wrist D435i is mounted on the gripper.}
\label{fig:hw}
\end{figure}

\paragraph{Tabletop Scene Description.}
The platform comprises a UR5e manipulator with a Robotiq 2F-85 parallel-jaw gripper mounted on an aluminium breadboard on its own stand, a separate table in front of it that carries the objects, and two Intel RealSense D435i cameras: one external RGB-D camera mounted on a tripod beside the table and one wrist camera rigidly attached to the end effector (\cref{fig:hw}).
A featureless black curtain behind the table bounds the external camera's view, so the static scene of the world anchor is the table top and the objects on it; in a base shift the table and the tripod move together, and the arm is the only part of the view whose position changes.
The two placements are complementary: the external camera provides global scene coverage and is the one that pose shifts affect; the wrist camera supplies near-field geometry at the moment of contact and returns to the training distribution once the start pose is homed.

\paragraph{Camera Description and Calibration.}
Both cameras stream $640\times480$ at 30\,Hz: the external camera aligned RGB-D, the wrist camera RGB only. The policy input is the external RGB resized to $224\times224$ and the wrist RGB, and the front end uses the external depth at full resolution.
The extrinsics of the external camera are obtained by hand--eye calibration once before data collection: a ChArUco board is fixed to the gripper, the arm visits 20 poses spread over the workspace, and the Tsai--Lenz solution~\cite{tsai1989hand} is refined by minimising the reprojection error (0.4\,px).
Together with the robot's base pose and one RGB-D frame of the empty table, the median of 30 frames with all movable objects removed, they form the calibration bundle of \method.
The wrist camera needs no calibration.

\paragraph{Realisation of Pose Shifts.}
\cref{tab:sup_poses} lists the shifts of the three conditions, the base shift as realised and the camera shifts as measured by re-calibration.
All poses are marked on the floor, and the table's feet and the tripod's feet are returned to the marks before every block of trials.

\begin{table}[t]
\centering
\footnotesize
\setlength{\tabcolsep}{4pt}
\caption{\textbf{Real-Robot Shift Poses}: translation and rotation of the camera relative to the table ($\dTc$) and of the arm's base relative to the table ($\dTb$); the base shift as realised, the camera shifts as measured by hand--eye re-calibration. Last column: the corresponding level of \bench.}
\label{tab:sup_poses}
\begin{tabular}{lccccc}
\toprule
 & \multicolumn{2}{c}{$\dTc$} & \multicolumn{2}{c}{$\dTb$} & \\
\cmidrule(lr){2-3}\cmidrule(lr){4-5}
Condition & cm & deg & cm & deg & Scale \\
\midrule
camera & 7.2 & 4.5 & 0 & 0 & L3 \\
base & 0 & 0 & 10.0 & 0 & L3 \\
both & 6.5 & 5.8 & 10.0 & 0 & L3 \\
\bottomrule
\end{tabular}
\end{table}

\paragraph{Object Description.}
We use everyday objects: a lemon among several toy fruits and a white basket; coloured blocks and cups of matching colours; a plush toy and a two-tier wire shelf; a slice of toast on a plate and a toaster.
The shelf and the toaster stay at the same spot across trials; all other objects are placed at the start of every trial from a pre-generated layout drawn uniformly within a task-specific region, and the same layouts are used for every method.
\section{Policy Learning Details}
\label{sup:learning}

\paragraph{Data Collection.}
Real-robot demonstrations are collected by teleoperation with a 3D SpaceMouse that commands end-effector velocities through the UR RTDE interface, with the operator standing outside the external camera's view so that the static background stays static.
Every demonstration starts from the same joint configuration, as \method's homing assumes.
We record at 10\,Hz the external RGB-D frame, the wrist RGB frame, the joint angles (needed by the arm anchor), the end-effector pose in the base frame and the gripper opening as the state, and the end-effector delta in the base frame (translation, rotation vector composed on the left, gripper command) as the action; all streams are time-stamped on the host and aligned to the 10\,Hz grid.
Failed demonstrations are discarded and idle frames removed, as in LIBERO, leaving 50 demonstrations per task.
The 50 initial layouts of each task are drawn by stratified sampling over the task's region and recorded in an interleaved order, so that the first 10 and the first 20 demonstrations, used for the few-shot policies, cover the region evenly.
Each dataset is checked by overlaying the forward-kinematics arm on the external images, by replaying recorded actions open-loop, and by running the front end on nominal frames, which must return the identity.
Simulation uses the public LIBERO demonstrations without modification.

\paragraph{Robot Cloud.}
The visual meshes of the links, the hand and the fingers are sampled with 4{,}000 surface points in total, placed at the link poses given by forward kinematics.
Points are transformed to the camera frame at the current hypothesis, projected with the pinhole intrinsics, and a model point is kept if it lands inside the image and its depth is within 3\,cm of the observed depth or in front of it.

\paragraph{ICP Schedule.}
Initial hypotheses are the prior $\Tbcz\hdTc$ composed with base yaw offsets of $\{-16,-8,0,8,16\}^\circ$ and planar offsets of $\{0,\pm 8\}$\,cm along each axis (25 starts); each runs six iterations of trimmed point-to-point ICP with a 10\,cm correspondence radius, and the best few continue for (10, 15, 10) iterations at (5, 2, 1)\,cm, keeping the best 80\% of correspondences.

\paragraph{Depth-Consistency Score.}
For a hypothesis $T$, the model cloud is projected and each covered pixel contributes $+1$ if $|D_{\mathrm{obs}}-D_{\mathrm{model}}|<1.5$\,cm and $-2$ if $D_{\mathrm{obs}}>D_{\mathrm{model}}+3$\,cm (the camera sees through the model); the hypothesis with the largest score is returned together with its residual.

\paragraph{World Anchor.}
SIFT with default parameters; keypoints within two pixels of a depth discontinuity larger than 3\,cm are discarded; matches are lifted to 3D and 5{,}000 RANSAC iterations with a 1\,cm inlier threshold fit a rigid transform by the Umeyama method.
Point-to-plane refinement uses 6{,}000 reference points with normals from central differences and 25 Gauss--Newton iterations with a 5\,mm Huber loss.
\section{Additional Results}
\label{sup:results}
\begin{figure*}[t]
\centering
\includegraphics[width=0.92\linewidth]{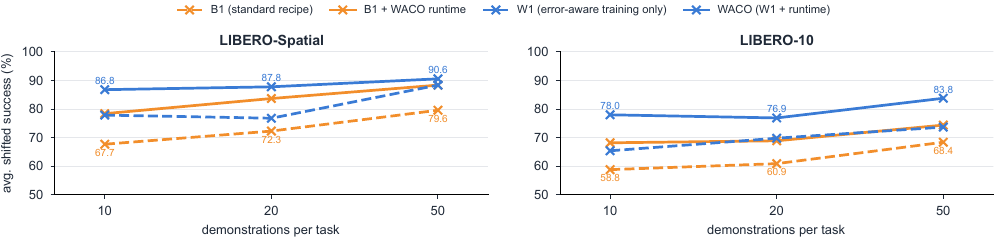}
\caption{\textbf{Few-Shot Scaling.} Average shifted success on \bench with the first 10, 20 and 50 demonstrations per task, same recipe and budget, per suite. The gap between \method and the standard recipe widens as data shrinks, because the pose variation that the recipe must absorb from data is removed analytically; both the runtime and the error-aware training contribute, and their combination is better than either.}
\label{fig:fewshot}
\end{figure*}

\cref{tab:lpfail} breaks the LIBERO-Plus results down by the state of the front end, \cref{fig:fewshot} shows the few-shot scaling per suite, \cref{tab:sup_persuite} lists the main comparison per suite and \cref{tab:sup_real} the per-task real-robot results behind \cref{tab:real}.

\paragraph{Where the Remaining Failures Are.}
\cref{tab:lpfail} joins a per-task probe of the front end over all 1{,}599 tasks with the episode outcomes.
Of the 101 failures of the first seed, 49 fall in the accurate-and-applied group and 38 of those are at the widest orbits, so most remaining failures are failures of the policy on heavily re-warped views, not of the estimator.
Where the front end has no usable estimate (67 tasks, almost all orbits of $60^\circ$ or more), \method runs the episode uncanonicalised and still succeeds more often than the public checkpoint, because the error-aware fine-tuning has seen comparable views.

\begin{table}[t]
\centering
\footnotesize
\caption{\textbf{LIBERO-Plus Outcomes by Front-End State} (corrected protocol, all 1{,}599 tasks; \method: mean over three evaluation seeds; no runtime: the public checkpoint without canonicalisation). \emph{Accurate}: within 2\,cm and $2^\circ$ of the true camera shift.}
\label{tab:lpfail}
\begin{tabular}{lccc}
\toprule
Front-end state & Tasks & \method & no runtime \\
\midrule
accurate, applied & 1{,}423 & 94.3 & 64.4 \\
inaccurate, applied & 95 & 76.1 & 65.3 \\
no usable estimate & 67 & 64.2 & 46.3 \\
correct but not trusted & 14 & 97.6 & 71.4 \\
\bottomrule
\end{tabular}
\end{table}

\begin{table}[t]
\centering
\footnotesize
\caption{\textbf{Per-Suite Results} of the main comparison (\cref{tab:main}); average over the nine shifted conditions (900 episodes per cell).}
\label{tab:sup_persuite}
\begin{tabular}{lccc}
\toprule
Method & Spatial & LIBERO-10 & Mean \\
\midrule
B1 & 79.6 & 68.4 & 74.0 \\
B1 + 2D aug & 83.6 & 71.4 & 77.5 \\
B2 & 88.4 & 73.7 & 81.1 \\
\method runtime on B1 & 88.4 & 74.4 & 81.4 \\
\rowcolor{oursblue}
\textbf{\method} & \textbf{90.6} & \textbf{83.8} & \textbf{87.2} \\
GT oracle on W1 & 91.0 & 86.3 & 88.7 \\
\bottomrule
\end{tabular}
\end{table}

\begin{table*}[t]
\centering
\footnotesize
\setlength{\tabcolsep}{5pt}
\caption{\textbf{Real-Robot Per-Task Results} (success rate in \%, 20 trials per task and condition). Mean rows are highlighted; $\dagger$ uses the re-calibrated ground-truth shifts.}
\label{tab:sup_real}
\begin{tabular}{llccccc}
\toprule
Method & Condition & Lemon to Basket & Color Matching & Toy on Shelf & Toast into Toaster & Mean \\
\midrule
\multirow{4}{*}{\pizf (B1)} & Nominal & 95 & 95 & 90 & 90 & 92.5 \\
 & Camera & 80 & 65 & 50 & 45 & 60.0 \\
 & Base & 60 & 45 & 30 & 25 & 40.0 \\
 & Both & 50 & 35 & 20 & 15 & 30.0 \\
\rowcolor{meanblue}
 & Mean (shifted) & 63.3 & 48.3 & 33.3 & 28.3 & 43.3 \\
\midrule
\multirow{4}{*}{B1 + 2D aug} & Nominal & 95 & 95 & 95 & 90 & 93.8 \\
 & Camera & 90 & 80 & 65 & 55 & 72.5 \\
 & Base & 65 & 50 & 30 & 25 & 42.5 \\
 & Both & 55 & 40 & 20 & 15 & 32.5 \\
\rowcolor{meanblue}
 & Mean (shifted) & 70.0 & 56.7 & 38.3 & 31.7 & 49.2 \\
\midrule
\multirow{4}{*}{\method runtime on B1} & Nominal & 95 & 90 & 90 & 90 & 91.3 \\
 & Camera & 90 & 80 & 70 & 60 & 75.0 \\
 & Base & 85 & 75 & 70 & 60 & 72.5 \\
 & Both & 80 & 70 & 55 & 45 & 62.5 \\
\rowcolor{meanblue}
 & Mean (shifted) & 85.0 & 75.0 & 65.0 & 55.0 & 70.0 \\
\midrule
\multirow{4}{*}{\textbf{\method}} & Nominal & \textbf{95} & \textbf{95} & \textbf{95} & \textbf{95} & \textbf{95.0} \\
 & Camera & \textbf{95} & \textbf{90} & \textbf{85} & \textbf{70} & \textbf{85.0} \\
 & Base & \textbf{90} & \textbf{85} & \textbf{80} & \textbf{65} & \textbf{80.0} \\
 & Both & \textbf{85} & \textbf{80} & \textbf{75} & \textbf{60} & \textbf{75.0} \\
\rowcolor{oursblue}
 & Mean (shifted) & \textbf{90.0} & \textbf{85.0} & \textbf{80.0} & \textbf{65.0} & \textbf{80.0}\,\gain{+36.7} \\
\midrule
\multirow{4}{*}{GT oracle on W1$^\dagger$} & Nominal & 95 & 95 & 95 & 95 & 95.0 \\
 & Camera & 95 & 90 & 90 & 75 & 87.5 \\
 & Base & 95 & 90 & 85 & 75 & 86.3 \\
 & Both & 90 & 85 & 80 & 75 & 82.5 \\
\rowcolor{meanblue}
 & Mean (shifted) & 93.3 & 88.3 & 85.0 & 75.0 & 85.4 \\
\bottomrule
\end{tabular}
\end{table*}
\section{Supplementary Video}
\label{sup:videos}
The supplementary video shows \method on all four real-robot tasks, two uncut trials per task: one in the nominal condition and one under the base shift.
Each clip is recorded from a third viewpoint that shows the arm, the table and the external camera on its tripod, so that the shift itself is visible.

\section{Visual Results}
\label{sup:qualitative}
\begin{figure*}[t]
\centering
\includegraphics[width=0.96\linewidth]{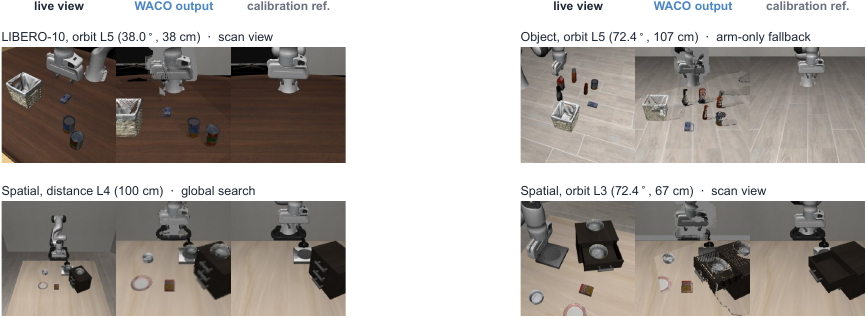}
\caption{\textbf{Visual Result.} LIBERO-Plus views far outside the training distribution: the live frame, the frame canonicalised by \method, and the calibration reference, with the true viewpoint change and the stage of the front end that produced the estimate. Pixels the live camera did not see are filled from the reference, so objects appear only where the live view observed them.}
\label{fig:qual}
\end{figure*}

\begin{figure*}[t]
\centering
\includegraphics[width=0.9\linewidth]{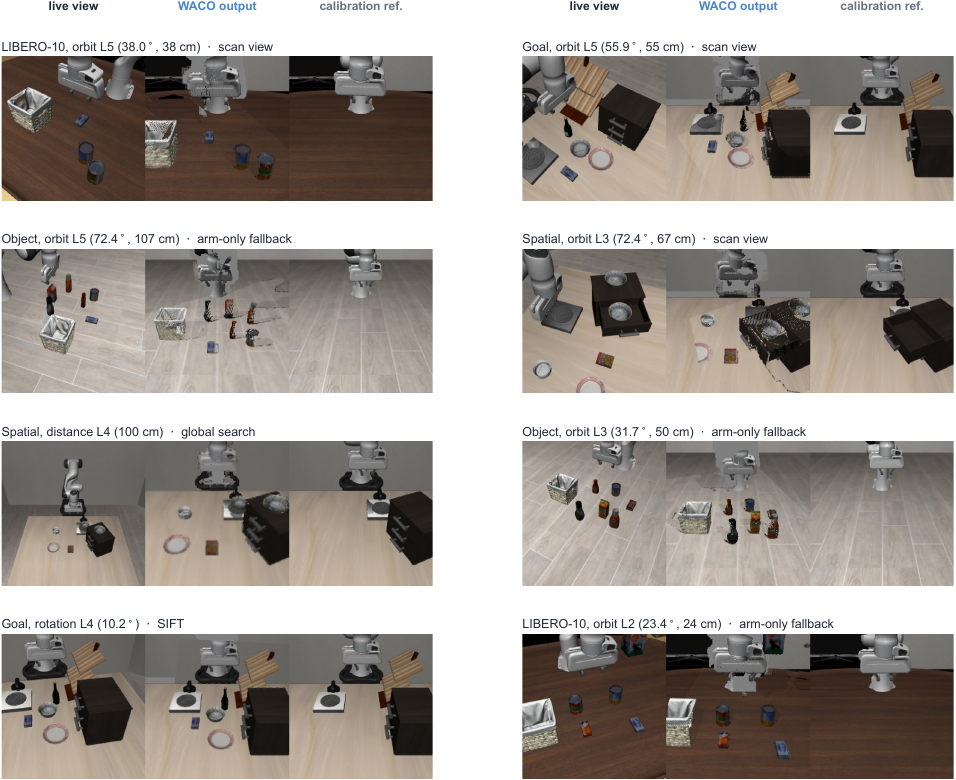}
\caption{\textbf{More Visual Results} on the LIBERO-Plus camera track: live view, the view canonicalised by \method, and the calibration reference.}
\label{fig:sup_qual}
\end{figure*}

\cref{fig:qual} and \cref{fig:sup_qual} show canonicalised third-person frames across the LIBERO-Plus camera track (live view, the view canonicalised by \method, and the calibration reference), with the true viewpoint change and the stage of the front end that produced the estimate. The last rows include the hardest cases, where the arm-only fallback is the only usable estimate.

\end{document}